\RequirePackage{fix-cm}
\documentclass[twocolumn]{svjour3}          % twocolumn
\smartqed  % flush right qed marks, e.g. at end of proof

\usepackage{graphicx}
\usepackage{bm}
\usepackage{subfig}
\usepackage{amssymb}
\usepackage{gensymb}
\usepackage{natbib}
\usepackage[pagebackref=true,breaklinks=true,letterpaper=true,colorlinks,bookmarks=false]{hyperref}

\journalname{International Journal of Computer Vision}
\begin{document}

\title{DenseReg: Fully Convolutional Dense Shape Regression In-the-Wild}

% \titlerunning{DenseReg}        % if too long for running head

\author{R{\i}za Alp G\"uler, Yuxiang Zhou, George Trigeorgis, Epameinondas Antonakos, Patrick Snape, Stefanos Zafeiriou, Iasonas Kokkinos
}

\authorrunning{G\"uler, Zhou, Trigeorgis, Antonakos, Snape, Zafeiriou, Kokkinos} % if too long for running head

\institute{
    R{\i}za Alp G\"uler
    \email{riza.guler@inria.fr}  
    Yuxiang Zhou 
    \email{yuxiang.zhou10@imperial.ac.uk}  
    \and 
    George Trigeorgis
    \email{gt108@imperial.ac.uk}  
    \and
    Epameinondas Antonakos
    \email{e.antonakos@imperial.ac.uk}  
    \and
    Patrick Snape
    \email{p.snape@imperial.ac.uk}  
    \and
    Stefanos Zafeiriou
    \email{s.zafeiriou@imperial.ac.uk}  
    \and
    Iasonas Kokkinos
    \email{i.kokkinos@cs.ucl.ac.uk}  
}

\date{Received: date / Accepted: date}

\maketitle

\begin{abstract} 
In this work we use deep learning to establish dense correspondences between a 3D object model and an image  ``in the wild''. We introduce `DenseReg', 
 a  fully-convolutional neural network (F-CNN) that  {\em{dens}}ely {\em{reg}}resses  at every foreground pixel a pair of U-V template coordinates in a single feedforward pass. 
To train DenseReg we construct a supervision signal by combining 3D deformable model fitting and 2D landmark annotations. We define the regression task in terms of the intrinsic, U-V coordinates of a 3D deformable model that is brought into correspondence with image instances at training time. A host of other object-related tasks (e.g. part segmentation, landmark localization) are shown to be by-products of this task, and to largely improve thanks to its introduction. 
We obtain highly-accurate regression results by combining ideas from semantic segmentation with regression networks, yielding a `quantized regression' architecture that first obtains a quantized estimate of position through classification, and refines it through regression of the residual.
We show that such networks can boost the performance of existing state-of-the-art systems for pose estimation. 
Firstly, we show that 
our system can serve as an initialization for Statistical Deformable Models, as well as an element of cascaded architectures that jointly localize landmarks and estimate dense correspondences. We also show that  the obtained dense correspondence can act as  a  source  of  ‘privileged  information’  that  complements and extends the pure landmark-level annotations, accelerating and improving the training of pose estimation networks.
We report state-of-the-art performance  on the challenging 300W benchmark for facial landmark localization and on the MPII and LSP datasets for human pose estimation.
DenseReg code and demonstrations are made available at \url{http://alpguler.com/DenseReg.html}.
\end{abstract}

% main file
\section{Introduction}
\label{sec:introduction}

\begin{figure*}[h!]
\begin{center}
   \includegraphics[width=1\linewidth]{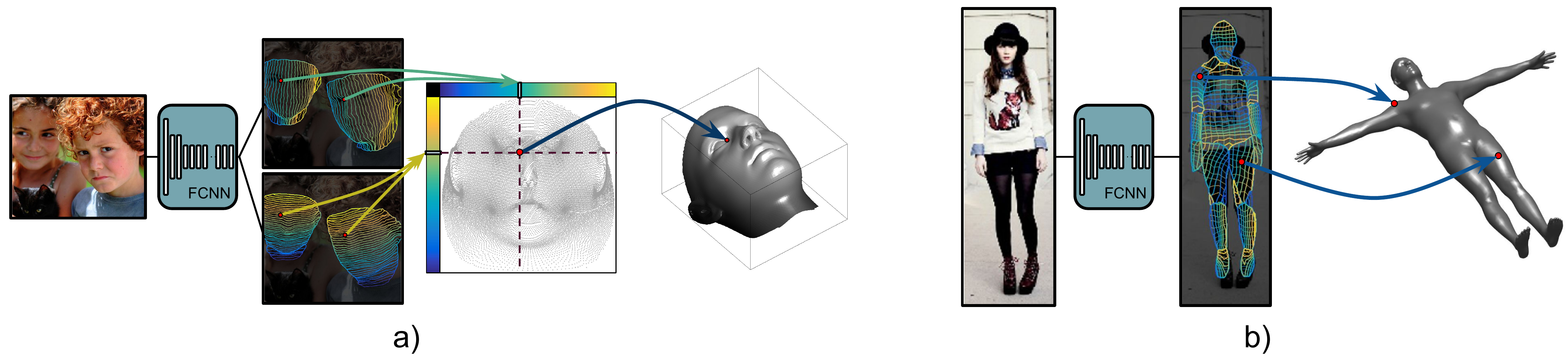}
\end{center}
   \caption{ We introduce a fully convolutional neural network that regresses from the image to a ``canonical'', deformation-free parameterization of the shape surface, effectively yielding a dense 2D-to-3D surface correspondence field. The system is depicted for dense correspondence between template shapes of \textit{a):} human face \textit{b):} human body.}
\label{fig:intro}
\end{figure*}

Deep Convolutional Neural Networks CNNs \citep{lecun1998gradient}
have revolutionized computer vision over the last decade, starting from  image classification \citep{krizhevsky2012imagenet, simonyan2014very, szegedy2015going,he2016deep}, and then moving on to tasks such as object detection \citep{girshick2014rich}, semantic segmentation \citep{long2015fully,chen2016deeplab} and pose estimation \citep{chen2014articulated, tompson2014joint,yang2016end,newell2016stacked}.
The order in which these tasks were successfully tackled can  be associated with the level of spatial detail at which the problem is addressed, starting from boxes, moving on to regions, and eventually getting to the  pixel-level labelling.

In this work we push further the envelope of tasks that can be addressed by CNNs, and consider a task that lies at the end of the `location detail' spectrum.
 Rather than characterizing  the region, or a few select points  that relate to an object, we aim at establishing a dense correspondence between 2D  and 3D surface coordinates, where the surface represents a template (or atlas) for a visual category, such as the human face or body. 
We show that  this task can be successfully addressed in an entirely feedforward manner by employing a discriminatively-trained CNN. 

%capture all of the excruciating details underlying image-to-model matching
%tackle the much more challenging task of establishing a 2D to 3D correspondence {\emph{in the wild}} by leveraging upon recent advances in semantic segmentation \citep{CP2015Semantic}. 
%one of the most `exchaustively detailed' visual tasks, namely  dense image-to-surface correspondence `in the wild',  
%semantic segmentation fine-grained categorization \citep{zhang2014part}, among others.
%Motivated by the gap between discriminatively trained systems for detection and category-level deformable models, we propose a system that combines the merits of both.
%Discriminative learning-based approaches typically pursue invariance to shape deformations, for instance by employing  local `max-pooling' operations to ellicit responses that are invariant to {\emph{local}} translations. As such, these models can reliably  detect patterns irrespective of their deformations through efficient, feedforward algorithms. At the same time, however, this discards useful shape-related information and only delivers a single categorical decision per position. 
%We introduce a discriminatively trained network to obtain, in a fully-convolutional manner, dense correspondences between an input image and a deformation-free template coordinate system. 

In order to accomplish this 
 we exploit the availability of manual landmark annotations ``in-the-wild'' in order to fit a 3D template; this provides us with a dense correspondence field, from the image domain to the 2-dimensional, $U-V$ parameterization of the surface. We then train a fully convolutional network that densely regresses from the image pixels to this $U-V$ coordinate space. This combines the fine-grained discrimative power of statistical deformable models with the ``in the wild'' operation of fully-convolutional neural networks. 
 We draw inspiration from recent successes of object detection at the task of bounding box regression \citep{ren2015faster}, and 
 introduce a method that blends classification and regression to accurately regress the 2D template coordinates of every foreground pixel.
 
 As we show experimentally, the proposed  feedforward architecture outperforms substantially more involved systems developped in particular for facial landmark localization while also outperforming the results of systems trained on lower-granularity tasks, such as facial part segmentation.
We can also seamlessly integrate this method with iterative, deformable model-based algorithms to obtain results that constitute the current state-of-the-art on large-scale, challenging facial landmark localization benchmarks.

%This provides us with dense and fine-grained correspondence information, as in the case of SDMs, while at the same time being independent of any initialization procedure, as in the case of discriminatively trained `fully-convolutional' networks. 

Furthermore, we show that by exploiting the established dense shape correspondence one can substantially improve the performance of CNNs trained for articulated body pose estimation and facial landmark localization and accelerate their training. 
In particular, recent CNN-based body and facial pose estimation works  only implicitly capture shape-based e.g. through cascading \citep{newell2016stacked}.  Instead, we further exploit shape for CNN  training by introducing an auxiliary dense correspondence supervision signal that acts like a source  of  `Privileged Information' \citep{VapnikV09,lopez2015unifying,ChenJFY17}.
Our experiments show that the cascading and dense supervision approaches are clearly complementary and can be combined, yielding faster and improved convergence. 

%\footnote{`Privileged Information' was proposed in \citep{VapnikV09} where it is argued that one can simplify training through the use of an `Intelligent Teacher' that in a way explains the supervision signal, rather than simply penalising misclassifications. This technique was recently used in deep learning for the task of image classification \citep{ChenJFY17}.}.

%We note that  this additional information is only available during training,  only serves as a means of simplifying the training problem, and only requires landmark-level annotations, as all current methods do.

%%%%%%%%%%%%%%%%%%%%%%%%%%%%%%%%%%%%%%%%

We can summarize our contributions as follows:
\begin{itemize}
\item We introduce the task of dense shape regression in the setting of CNNs, and exploit the notion of a deformation-free UV-space to construct target ground-truth signals (Sec.\ref{sec:SDMs}).
\item We propose a carefully-designed fully-convolutional shape regression system that exploits ideas from semantic segmentation and dense regression networks. Our \textit{quantized regression} architecture~(Sec.\ref{sec:quantized}) is shown to substantially outperform simpler baselines that consider the task as a plain regression problem. 
\item We use dense shape regression to jointly tackle a multitude of problems, such as landmark localization or semantic segmentation.
In particular, the template coordinates allow us to transfer to an image multiple annotations constructed on a single template system, and thereby tackle multiple problems through a single network.
\item We use the regressed shape coordinates for the initialization of statistical deformable models; systematic evaluations on facial analysis benchmarks show that this yields substantial performance improvements  on tasks.
\item We  show that a cascaded architecture that jointly regresses dense correspondences and sparse landmarks leads to improved localization in both articulated body pose estimation and facial landmark localization. 
\item We  demonstrate the generic nature of the method by applying it to the task of estimating dense correspondence in other object, such as the human ear.
\end{itemize}
A preliminary version of the paper has appeared in CVPR 2017 \citep{guler2016densereg}. The present version bears substantial novelties and extended experiments. The most important novelty is the design of end-to-end deep networks for joint dense shape correspondence estimation and articulated body pose estimation, where we demonstrate that dense correspondence largely improves the performance of articulated pose estimation.
 
%\item the design of end-to-end deep networks for joint dense shape correspondence estimation and facial landmark %localisation and tracking.  
%\end{itemize}

The rest of the manuscript is summarized as follows: In Section \ref{sec:SDMs} we present the idea of establishing dense correspondences between the normalized 3D model space of a deformable object and 2D images. In Section \ref{sec:quantized} we present a deep learning framework for establishing dense correspondences using Deep Convolutional Neural Networks (DCNNs) and in particular a quantized regression approach tailored to the task. In the same section we also present DCNN frameworks for joint articulated pose estimation and dense shape correspondence estimation. We present experiments in Section \ref{sec:experiments} and conclude in Section \ref{S:CONCLUSIONS}.

% Our system is particularly simple to implement, as it relies on a variation of the broadly adopted 
% Deeplab system \citep{CP2015Semantic}.

\section{Previous work}

Our work draws inspiration from two threads of research:  Convolutional Neural Networks (CNNs) and Statistical Deformable Models (SDMs).
Our starting point is the understanding that 
planar object deformations, e.g. due to pose or expression, result in challenging but also informative signal variations. While CNNs are typically geared towards discounting
the effects of deformations,  SDMs aim at capturing their details; our work aims
at capitalizing on the power of both approaches.

%A common theme in these works is that DCNNs trained in an end-to-end manner  deliver  strikingly better results than systems relying on carefully engineered features, such as SIFT or HOG features.
%This success can be partially attributed to the built-in  invariance of DCNNs to local image transformations, which underpins their ability to learn hierarchical abstractions of data \citep{zeiler2014visualizing}. While this invariance is clearly desirable for high-level vision tasks, it can hamper low-level tasks, such as pose estimation  and semantic segmentation - where we want precise localization, rather than abstraction of spatial details.  %% As 

In particular, several recent works in deep learning have aimed at enriching deep networks with information about shape by explicitly modelling {\em the effect} of  similarity transformations \citep{PapandreouKS15}
or non-rigid deformations \citep{JaderbergSZK15,HandaBPSMD16,ChenHW016}; several of these have found success in classification \citep{PapandreouKS15}, fine-grained recognition  \citep{JaderbergSZK15}, and also face detection \citep{ChenHW016}. There are works \citep{lades1993distortion,pedersoli2015elastic} that model the deformation via optimization procedures, whereas we obtain it in a feedforward manner and in a single shot. In these works, shape is treated as a nuisance, while we treat it as the goal in itself. 
Earlier discriminatively trained models exploited depth data for 3D human body correspondence \citep{TaylorSSF12,WeiHCVL15}, while recent works on 3D surface correspondence \citep{Br1,Br2} have shown the merit of CNN-based unary terms for correspondence. Instead our work relies entirely on RGB inputs.
Moving beyond discriminatively training, recent work \citep{ThewlisBV17a}  has explored how CNNs can be used for unsupervised non-rigid alignment of images, along the lines of earlier works on congealing \citep{Learned-Miller06,KokkinosY07}. Even though certainly promising, the results are still not directly comparable with the present state-of-the-art on challenging benchmarks.

Approaches that rely on  Statistical Deformabe Models (SDMs), such as Active Appearance Models (AAMs) or 3D Morphable Models (3DMMs) aim at explicitly recovering  dense correspondences between a deformation-free template and the observed image, rather than trying to discard them. 
This allows to both represent shape-related information (\textit{e.g.} for facial expression analysis) and also to obtain invariant decisions after registration (\textit{e.g.} for identification). 
Explicitly representing shape  can have substantial performance benefits, as is witnessed in the majority of facial analysis tasks requiring detailed face information e.g. identification \citep{TaigmanYRW14}, landmark localisation \citep{sagonas2016300}, 3D pose estimation, as well as 3D face reconstruction ``in-the-wild'' \citep{jourabloo2016large}. 
However alignment-based methods are limited in two respects. Firstly they require an initialization from external systems, which can become increasingly challenging for elaborate SDMs: both AAMs and 3DMMs require at least a bounding box as initialization and 3DMMs may further require position of specific facial landmarks. Furthermore, the problem of fitting a 3DMM of human body is even more challenging requiring further assumptions \citep{lassner2017unite}. In general, SDM fitting  requires iterative, time-demanding optimization algorithms, especially when the initialisation is far from the solution \citep{booth20173d}. The advent of Deep Learning has made it possible to replace the iterative optimization task with iterative regression problems \citep{trigeorgis2016mnemonic}, but this does not alleviate the need for initialization and multiple iterations. 

Bridges between the detection and SDM-based approaches have often been pursued in the past.
Shape information has commonly been used in pose estimation in the form of a prior on the relative positions of parts, following the Pictorial Structures model\citep{fischler1973representation}. This is explicitly represented in the form of energy terms in deformable part-based methods
(DPMs) ~\citep{felzenszwalb2008discriminatively} and related probabilistic graphical model (PGM) approaches to the problem of pose estimation ~\citep{andriluka2009pictorial,sapp2010adaptive,yang2011articulated,sapp2013modec} and face landmark localization \citep{zhu2012face}. Even though CNNs were originally only used as parts of graphical model in \citep{jain2013learning,tompson2014joint,chen2014articulated,yang2016end},  more recently CNN-based architectures have been shown able to exploit shape and context implicitly via  stacked and  learnable part localization operations, through both non-recurrent \citep{wei2016convolutional,bulat2016human,newell2016stacked} and recurrent \citep{belagiannis2016recurrent} refinements. 

Instead, as we show in the following,  the present work shows that a feedforward CNN can jointly deliver detection and landmark localization by the introduction of an appropriate supervision signal, the introduction of a customized regression architecture, and combining dense supervision with modern cascaded architectures. 

\section{From Statistical Deformable Models to Network Supervision}\label{sec:SDMs}
Following the deformable template paradigm \citep{yuille1991deformable,Grenander1991}, we consider that object instances are obtained by deforming a prototypical object, or `template', through  dense deformation fields. 
This makes it possible  to factor  object variability within a category into variations that are associated to  deformations, generally linked to the object's 2D/3D shape, and variations that are associated to appearance (or, `texture' in graphics), e.g. due to facial hair, skin color, or illumination. 

This factorization largely simplifies the  modelling task. SDMs use it as a stepping stone for the construction of parametric models of deformation and appearance. For instance, in AAMs a combination of Procrustes Analysis, Thin-Plate Spline warping and PCA is the standard pipeline for learning a low-dimensional linear subspace that captures category-specific shape variability \citep{cootes2001active}. Even though we have a common starting point, rather than trying to construct a linear generative model of deformations, we treat the image-to-template correspondence as a vector field that our network tries to regress.

In particular, we start from a template \\
$\bm{X} = [\bm{x}_1^\top ,\bm{x}_2^\top,...\bm{x}_m^\top]^\top \in \mathbb{R}$, where each $\bm{x}_j \in \mathbb{R}^3$ is a vertex location of the mesh in 3D space. 
This template could be any facial mesh, but in practice it is most useful to use a topology that is in correspondence with a 3D statistical shape model such as \citep{booth3d2} or \citep{paysan20093d}.
We compute a bijective mapping $\psi$, from template mesh $\bm{X}$ to the 2D canonical space $\bm{U} \in \mathbb{R}^{2\times m}$, such that  
\begin{equation}
\psi(\bm{x}_j) \mapsto \bm{u}_j \in \bm{U}  \quad  ,  \quad  \psi^{-1}(\bm{u}_j) \mapsto \bm{x}_j .
\end{equation} 

\subsection{Supervision for the face template}

We exploit the availability of facial landmark annotations ``in the wild'', to fit the template face to the image by obtaining a coordinate transformation for each vertex $\bm{x}_j$. 
We use the fittings provided by \citep{zhu2016face} which were fit using a modified 3DMM implementation \citep{romdhani2005estimating}. However, for the purpose of this paper, we require a per-pixel estimate of the location in UV space on our template mesh and thus do not require an estimate of the projection or model parameters as required by other 3D landmark recovery methods \citep{jourabloo2016large,zhu2016face}. The per-pixel UV coordinates are obtained through rasterization of the fitted mesh and non-visible vertices are culled via z-buffering.

\begin{figure}
\centering
\includegraphics[trim={7.2cm 18.3cm 2.7cm 6.5cm}, clip, width=\linewidth]{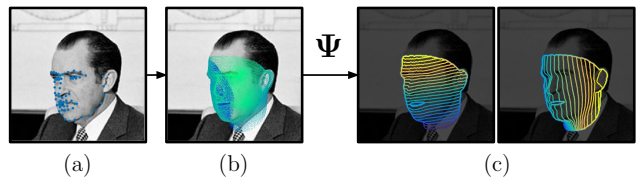}
\caption{Ground-truth generation: \emph{(a)} Annotated landmarks. \emph{(b)} Template shape morphed based on the landmarks. \emph{(c)} Deformation-free coordinates (${u^h}$ and ${u^v}$), obtained by unwrapping the template shape, transferred to image domain. }

\label{fig:GT}
\end{figure}

The mapping $\psi$ is obtained via the cylindrical unwrapping described in \citep{booth2014optimal}. Thanks to the cylindrical unwrapping, we can interpret these coordinates as being the horizontal and vertical coordinates while moving on the face surface: ${u}_j^h \in [0,1]$ and ${u}_j^v \in [0,1]$. Note that this semantically meaningful parameterization has no effect on the operation of our method.
As  illustrated in \ref{fig:GT}, once the transformation from the template face vertices to the morphed vertices is established, the  $\bm{u}_j$ coordinates of each visible vertex on the canonical face can be transferred to the image space. This establishes the ground truth signal for our subsequent regression task.

\subsection{Supervision for the human body template}

\begin{figure}[h!]
\begin{center}
   \includegraphics[width=1 \linewidth ]{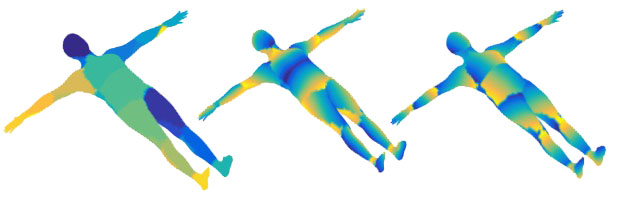}
\end{center}
   \caption{ \textit{Top:} Index, U and V fields displayed on the SMPL model. \textit{Bottom:} Dense correspondence results presented as input image fused with estimated UV coordinates, estimated UV coordinates and groundtruth UV coordinates respectively. A customized colour-coding is used for a clear demonstration of correspondence.}
\label{fig:IUV}
\end{figure}

We use the recently proposed "Unite the People" (UP) dataset~\citep{lassner2017unite}, which provides a 3D deformable human shape model \citep{loper2015smpl} in correspondence with images from LSP \citep{Johnson10}, MPII~\citep{andriluka14cvpr}, and FashionPose~\citep{dantone2013human} datasets. The dataset is obtained by solving an optimization problem of~\citep{bogo2016keep} to fit the surface given annotated landmarks and manually obtained segmentation masks for human bodies. The fits are filtered through crowdsourcing by manual elimination bad samples resulting into a total of 8515 images.
In order to handle the complex geometry of the human shape, we manually partition the surface into 25 patches each of which is isomorphic to the plane. Each vertex on the mesh is assigned a patch label, $I$. We establish a deformation-free coordinate system for each patch by applying multidimensional-scaling to corresponding vertices. This is followed by a normalization to obtain fields $U,V \in [0,1]$.  The $I$ ,$U$ and $V$ fields on the SMPL model\citep{loper2015smpl} is presented in Fig.~\ref{fig:IUV}.

%%%%%%%%%%%%%%%%%%%%%%%%%%%%%%%%%%%%%%%%%%%%%%%%%%%%%%%%%%%%%%%%%%%%%%%%%%%%%%%%%%%%%%%%%%%%%%%%%%%%%%%%%%%%%%%%%%%%%%%%%%%%%%%%%%%%%%%%%

\section{Quantized Regression}\label{sec:quantized}
\begin{figure*}[t]
\centering
\includegraphics[trim={5.2cm 20cm 3.7cm 4.5cm}, clip, width=0.95\linewidth ]{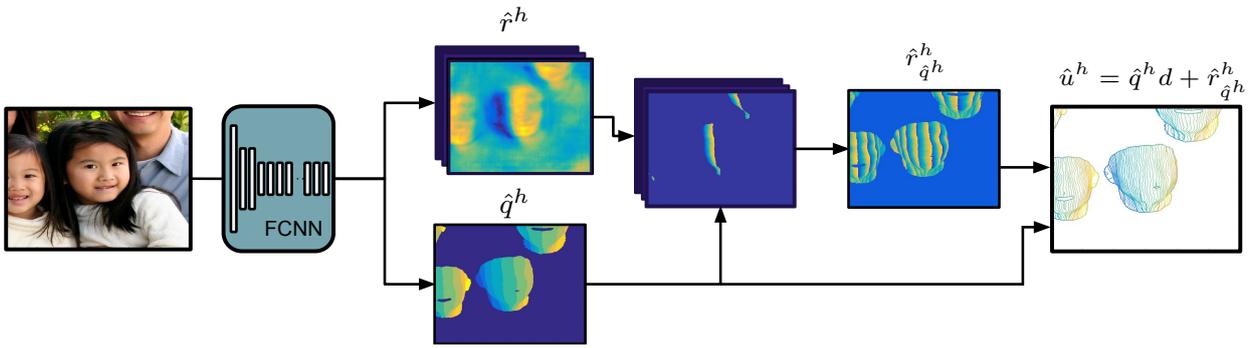}
\caption{Proposed Quantized Regression Approach for the horizontal correspondence signal: The continuous  signal is regressed by first estimating a grossly quantized (or, discretized) function  through a classification branch. For each quantized value $\hat{q}^h$ we use a separate residual regression unit's prediction, $\hat{r}^h_{\hat{q}^h}$, effectively multiplexing the different residual predictions. These are added to the quantized prediction, yielding a smooth and accurate correspondence field. }
\vspace{-0.35cm}
\label{fig:Pipeline}
\end{figure*}

Having described how we establish our supervision signal, we now turn to the task of estimating it through a convolutional neural network (CNN). 
Our aim is to estimate at any image pixel that belongs to a face region the values of  $\bm{u} =[u^h, u^v]$. We need to also identify non-face pixels,  e.g. by predicting a `dummy' output. 

One can phrase this problem as a generic regression task and attack it with the powerful machinery of CNNs. Unfortunately, the best performance that we could obtain this way was quite underwhelming, apparently due to the task's complexity. Our approach is to quantize and estimate the quantization error separately for each quantized value. Instead of directly regressing $u$, the quantized regression approach lets us solve a set of easier sub-problems, yielding improved regression results.

In particular,	instead of  using a CNN as a `black box' regressor, we draw inspiration from the success of recent works on semantic part  segmentation \citep{tsogkas2015deep,CP2016Deeplab}, and landmark classification \citep{bulat2016human,bulat2016two}. These works have shown that CNNs can deliver remarkably accurate predictions when trained to predict \textit{categorical variables}, indicating for instance the facial part or landmark corresponding to each pixel. 
	
Building on these successes, we propose a hybrid method that combines a classification with a regression problem. Intuitively, we first identify a coarser face region that can contain each pixel, and then obtain a refined, region-specific prediction of the pixel's $U-V$  field. As we will describe below, this yields substantial gains in performance when compared to the baseline of a generic regression system. 

For the human bodies, the regions are modeled by hand and for the facial regions, we use a simple geometric approach:
We tesselate the template's surface with a cartesian grid, by uniformly and separately quantizing the $u^h$ and $u^v$ coordinates into $K$ bins, where $K$ is a design parameter. For any image that is brought into correspondence with the template domain, this induces a discrete labelling, which can be recovered by training a  CNN for classification.
	
\begin{figure}[h]
\begin{center}
   \includegraphics[width=1\linewidth ]{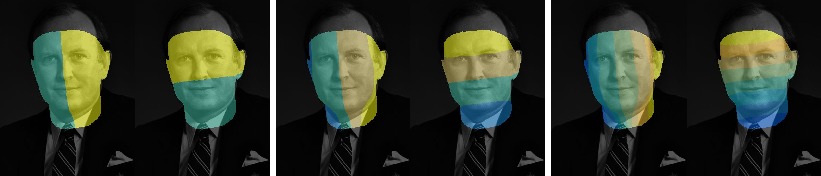}
\end{center}
   \caption{Horizontal and vertical tesselations obtained using $K=2,4$ and $8$ bins.}
   \vspace{-0.5cm}
\label{fig:DiscreteFaces}
\end{figure}

On Fig.~\ref{fig:DiscreteFaces}, the tesselations of different granularities are visualized. For a sufficiently large value of $K$ even a plain classification result could provide a reasonable estimate of the pixel's correspondence field, albeit with some staircasing effects. The challenge here is that as the granularity of these discrete labels becomes increasingly large, the amount of available training data decreases and label complexity increases. A more detailed analysis on the effect of label-space granularity to segmentation performance is provided in supplementary materials.

We propose to combine powerful classification results with a regression problem that will yield a  refined  correspondence estimate. For this, we compute the residual between the desired and quantized $U-V$  coordinates and add a separate module that tries to regress it. We train a separate regressor per facial region, and at any pixel only penalize the regressor loss for the responsible face region. We can interpret this form as a `hard' version of a mixture of regression experts \citep{JordanJ94}. 

The horizontal and vertical components $u^h,u^v$ of the correspondence field are predicted separately. This results in a substantial reduction in computational and sample complexity -  For $K$ distinct U and V bins we have $K^2$ regions; the classification is obtained by combining 2 $K$-way classifiers. Similarily, the regression mapping involves $K^2$ regions, but only uses $2 K$ one-dimensional regression units. The pipeline for quantized face shape regression is provided in Fig.~\ref{fig:Pipeline}.

We now detail the training and testing of this network;  for simplicity we only describe the horizontal component of the mapping. 
From the ground truth construction, every position $\bm{x}$ is associated with a scalar ground-truth value $u^h$. Rather than trying to predict $u^h$ as is, we transform it into a pair of discrete $q^h$ and continuous $r^h$ values, encoding the quantization and residual respectively:
\begin{equation} 
q^h =  \lfloor {\frac{u^h}{d}} \rfloor, \quad  r_i^h =   \left(u^h_i - q^h_i d  \right),
\end{equation}
where $d = \frac{1}{K}$ is the quantization step size (we consider $u^h,u^v$ coordinates to lie in $[0,1$]).

Given a common CNN trunk, we use two classification branches to predict $q^h, q^v$ and two regression branches to predict $r^h,r^v$ as convolution layers with kernel size $1\times1$. As mentioned earlier, we employ separate regression functions per region, which means that at any position we have $K$ estimates of the horizontal residual vector, $\hat{r}^h_{i},~i=1,\ldots,K$.

At test time, we let the network predict the discrete bin $\hat{q}^h$ associated with every input position, and then use the respective regressor output $\hat{r}^h_{\hat{q}^h}$ to obtain an estimate of $u$:
\begin{equation}   
\hat{u}^h =  \hat{q}^h d + \hat{r}^h_{\hat{q}^h}
\end{equation}

For the $q^h$ and $q^v$, which are modeled as categorical distributions,  we use  softmax followed by the cross entropy loss. For estimating $\hat{r}^h$ and $\hat{r}^v$, we use a normalized version of the smooth $L_1$ loss~\citep{girshick2015fast}. The normalization is obtained by dividing the loss by the number of pixels that contribute to the loss.

\subsection{Quantized Regression as Mixture of Experts}
\label{sec:MOE}

In our formulation,  $\hat{q}^h$ is modeled using a categorical distribution and is trained using softmax followed by cross entropy loss. This reconstruction can also be seen as:
\begin{equation}
\hat{u}^h = \sum_{i=0}^{K-1}  1_{(\hat{q}^h=i)}  (  i \cdot d + \hat{r}^h_{i}),
\end{equation}
where $(  i \cdot d + \hat{r}^h_{i})$ is the reconstruction by the $i_{\mathrm{th}}$ regressor and $1_{(\hat{q}^h=i)}$ is an indicator function, determining when the $i_{\mathrm{th}}$ regressor is active. Note that $i \cdot d$ is the value of $\hat{q}^h$, where $i_{\mathrm{th}}$ regressor is active.

Instead of this hard quantization, one can use a soft-quantization using the softmax function as:
\begin{equation}
\hat{u}^h = \sum_{i=0}^{K-1}  \bigg( \frac{e^{f^{q^h}_i}}{\sum_j e^{f^{q^h}_j}}   \bigg)  ( i \cdot d + \hat{r}^h_{i}),
\end{equation}
where $f^{q^h}$ is the output of the CNN branch trained for the quantized ($\hat{q}^h$) field. Notice that this is the \textit{mixture of experts} model, \cite{JordanJ94}, where the soft-quantization is analogous to the output of the gating network. It is straightforward to change our model accordingly: shifting each $\hat{r}^h_{i}$ by adding ($i \cdot d$) to the bias terms of the corresponding $1\times1$  convolutional layer and weighting each 'locally trained regressor' output by the softmax function and summing up. Since the parameters of the adapted network are not exactly optimized for this new soft-quantized model, we resort to end-to-end training. 

After the fine-tuning, the mixture of experts model performs as well as the quantized regression. Since no significant improvement in regression performance is observed, we have not performed any experiments related to facial analysis with this architecture. We consider that this differentiable representation could be more useful for instance as a spatial transformer network \cite{jaderberg2015spatial}, where the deformation field needs to be differentiable.

\subsection{Effect of Quantization to Regression Performance}
\label{sec: quantization_perf}

\begin{figure}[t]
    \centering
    \begin{minipage}{.7\linewidth}
        \centering
        \includegraphics[trim={0.5cm 0.5cm 0.7cm 0.5cm},clip,width=0.9\textwidth]{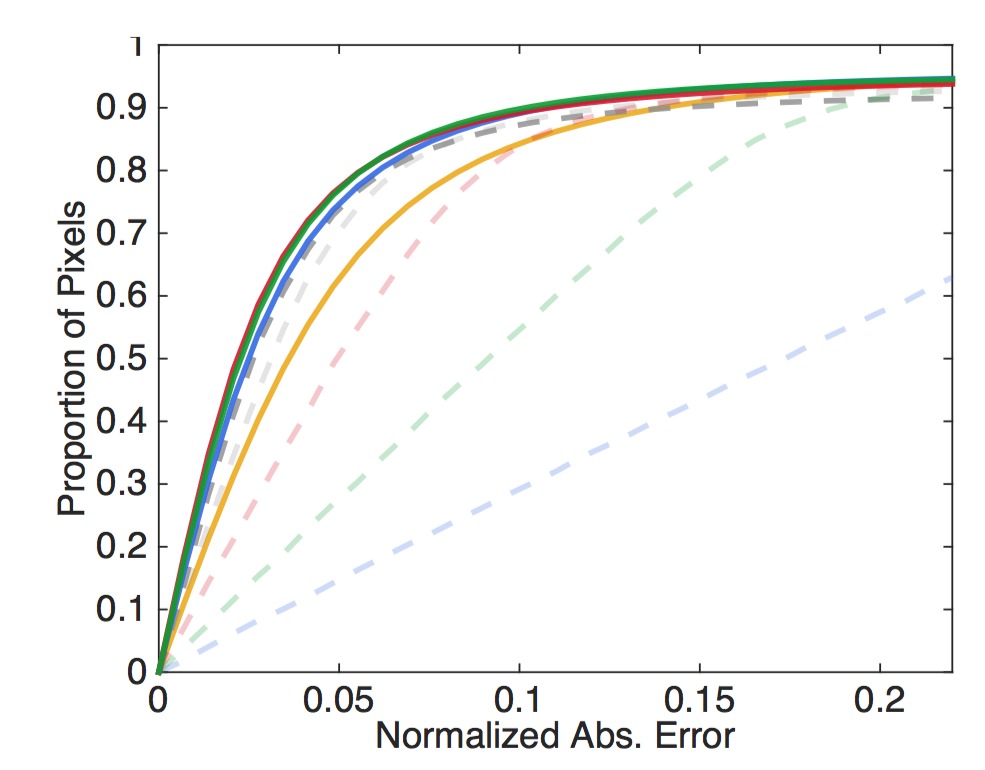}
    \end{minipage}%
    \begin{minipage}{0.3\linewidth}
        \centering
        \includegraphics[trim={1.5cm 1cm 10cm 1cm},clip,width=1\textwidth]{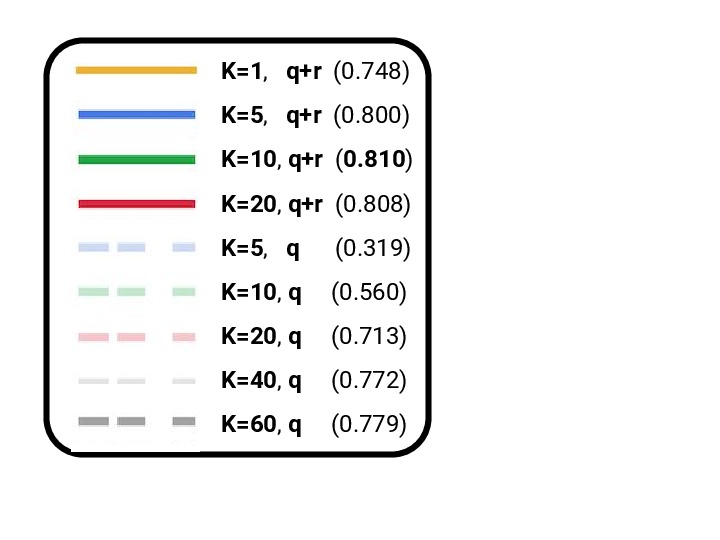}
    \end{minipage}
    \vspace{-0.35cm}
    \caption{Performance of $q$ and $r$, branches for various tesselation granularities of the human face, $K$. Areas under the curve(AUC) are reported.}
    \vspace{-0.15cm}
    \label{fig:exp}
    \vspace{-0.15cm}

\end{figure}

Compared to plain regression of the coordinates, the proposed quantized regression method achieves much better results. 
In Fig.\ref{fig:exp} we report results of an experiment that evaluates the contribution of the q-r branches separately for different granularities. The results for the quantized branch are evaluated by transforming the discrete horzintal/vertical label into the center of the region corresponding to the quantized horizontal/vertical value respectively.  The results  show the merit of adopting the classification branch, as the finely quantized results(K=40,60) yield  better coordinate estimates with respect to the non-quantized alternative {(K=1)}. After K=40, we observe an increase in the failure rate for the quantized branch. The experiment reveals that the proposed quantized regression outperforms both \textit{non-quantized} and the best of \textit{only-quantized} alternatives. For the human shape, the partitioning can be considered as the quantization.

\subsection{Supervisory Signals for Faces and Bodies}
\label{sec: supervisory_signal}

Different objects have different degrees of articulation. Hence, we have used different supervisory UV maps for faces and bodies. In particular, we found that it is sufficient to use as supervisory signal for faces two channels one of the U and one for the V coordinate and following the simple tessellation strategy defined Fig. \ref{fig:DiscreteFaces}. The network takes as input the three RGB channels and outputs three channels (one for the U coordinates, one for the V coordinates and one for tessellated coordinates). The training data have been produced by fitting a 3DMM that could describe both the identity, as well as the expression of a human face in "in-the-wild" images (see experimental result section for more details). 

On the other hand because body is a highly articulated object, with each part having each own self-occlusion maps (i.e., a hand or a foot can be occluded with the rest of the body being visible), we created a UV map per part. In total we split the body in 25 parts, as visualized in in Fig. \ref{fig:IUV}, and we applied quantised regression for each of the UV maps of the 25 body regions (parts). Hence, for the human body the network takes as input the three RGB channels of the image and outputs 75 channels. That is, three channels for each of the 25 parts (one for the U coordinates of the part, one for the V coordinates of the part and one for tessellated coordinates). 

\begin{figure*}[h!]
\begin{center}
   \includegraphics[width=0.49 \linewidth ]{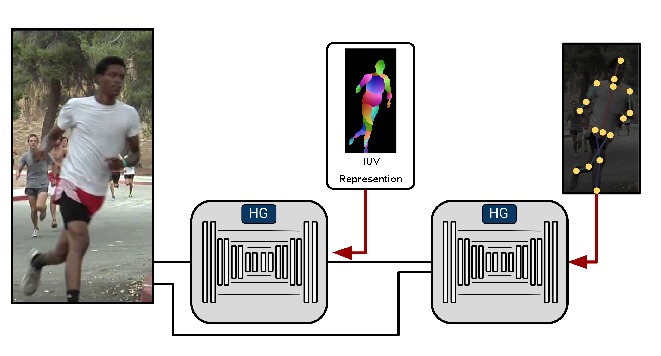}
   \includegraphics[width=0.49 \linewidth ]{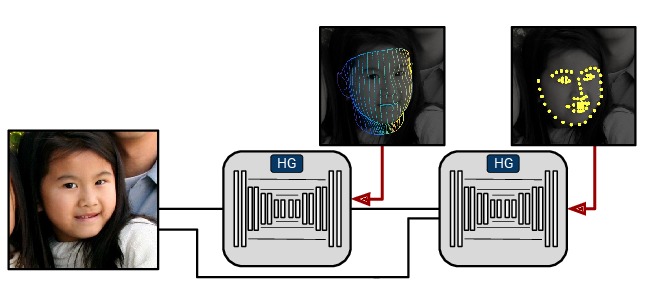}
\end{center}
   \caption{DenseReg cascade architecture for joint articulated pose estimation (body to the left, face to the right) and dense shape regression, wheredense correspondence supervision is obtained 3D Morphable Model fitting. Losses are shown in red. }
\label{fig:cascadeBody}
\end{figure*}

\subsection{A DenseReg Cascade for end-to-end dense shape regression and articulated pose estimation}
\label{sec: densereg_cascade}

Current algorithms for landmark localization and human pose estimation commonly address the learning problem in the form of a multi-class classification task, where each landmark defines its own class and the remainder of the image is labelled as background. Even though simple and effective, this training strategy provides a particularly sparse positive supervision signal, which asks a CNN to call everything other than a particular landmark a negative. We can intuitively say that our method simplifies this training `riddle', by providing information about dense correspondence between two surfaces. This fits  with the `Privileged Information' paradigm of  \citep{VapnikV09} where an `Intelligent Teacher' provides additional information during training that helps `understand' why a given  decision is taken. As a simple example, classifying a pixel as being a `knee' landmark can potentially be assisted by having dense correspondence maps, that  help the network solve the problem in a coarse-to-fine manner. Rather than rely on semantic supervision signals  we  rely on dense shape-level supervision.
 
Hence, motivated from the above we propose end-to-end trainable cascaded architectures which estimate dense correspondences and then these are used to improve articulated pose estimation. The architecture, coined DenseReg cascade, is depicted in Fig.~\ref{fig:cascadeBody}. In this particular, architecture the first network (which is in a form of hourglass) is used for dense shape regression. The output of the dense shape regression network is passed as privileged information in the second network which performs articulated body/face pose estimation.

%%%%%%%%%%%%%%%%%%%%%%%%
%%%%% EXPERIMENTS %%%%%%
%%%%%%%%%%%%%%%%%%%%%%%%
\section{Experiments}
\label{sec:experiments}
Herein, we evaluate the performance of the proposed method (referred to as \texttt{DenseReg}) on various face-related tasks. 
%The experiments are performed on challenging publicly available benchmarks and we report substantial improvements over the current state-of-the-art. 
In the following sections, we first describe the training setup (Sec.~\ref{sec:training_setup}) and then present extensive quantitative results on \emph{(i)}~semantic segmentation (Sec.~\ref{sec:exp_semantic_segmentation}), \emph{(ii)}~landmark localization on static images (Sec.~\ref{sec:exp_landmark_localization}), \emph{(iii)}~deformable tracking (Sec.~\ref{sec:exp_deformable_tracking}), \emph{(iv)}~ dense correspondence on human bodies (Sec.~\ref{sec:exp_human}), and \emph{(v)}~human ear landmark localization (Sec.~\ref{sec:exp_ear}).
%Finally, we report qualitative results on \emph{(v)}~depth estimation (Sec.~\ref{sec:exp_depth}) and \emph{(vi)}~3D shape estimation (Sec.~\ref{sec:exp_mm}), for which it is not possible to provide quantitative evaluation due to lack of established benchmarking. 
%We note that DenseReg is not trained or fine-tuned specifically for each one of those tasks. Its predictions are  used out-of-the-box the different task-specific evaluations. 
Due to space constraints, we refer to the supplementary material for additional qualitative results, experiments on monocular depth estimation and further analysis of experimental results.

%%%%%%%%%%%%%%%%%%%%%%%%%%%%%%%
%%%%%%% TRAINING SETUP %%%%%%%%
%%%%%%%%%%%%%%%%%%%%%%%%%%%%%%%
\subsection{Training Setup}
\label{sec:training_setup}

\textbf{Training Databases for Faces.} We train our system using the 3DDFA data of \citep{zhu2016face}. The 3DDFA data provides projection and 3DMM model parameters for the Basel \citep{paysan20093d} + FaceWarehouse \citep{cao2014facewarehouse} model for each image of the 300W database. We use the topology defined by this model to define our UV space and rasterize the images to obtain per-pixel ground truth UV coordinates.  Our training set consists of the LFPW trainset, Helen trainset and AFW, thus 3148 images
that are captured under completely unconstrained conditions
and exhibit large variations in pose, expression, illumination,
age, etc.
 Many of these images contain multiple faces, some of which are not annotated. We deal with this issue by employing the out-of-the-box DPM face detector of Mathias et al.~\citep{mathias2014face} to obtain the regions that contain a face for all of the images. The detected regions that do not overlap with the ground truth landmarks do not contribute to the loss. For training and testing, we have rescaled the images such that their largest side is 800 pixels.
 
 \textbf{Training Databases for Bodies}
 
 In order to create the supervisor UV signals for the body we made use of the recently proposed \textit{Unite the People} (UP)~\citep{lassner2017unite} dataset. The data is formed by automatically fitting the SMPL 3D model~\citep{loper2015smpl} (which has components that describe both the shape and the articulation of the human body in dense 3D). 
 For our experiments on human pose estimation with the cascaded architecture, in order to have a dense supervisory signal for all of the MPII and LSP images, we used the code provided by \citep{lassner2017unite}, to fit the SMPL model by minimizing the energy function proposed in ~\citep{bogo2016keep}. Even though some fits are erroneous, which are filtered in the UP dataset, the incorporation of dense correspondences yield improved pose estimation results. We used the estimated 3D shape and camera parameters to render a pixel-vertex correspondence map, where each image pixel is labeled with the corresponding vertex index and compute the corresponding UV maps. 
 
 The pose estimation experiments were performed on two well known body pose databases: MPII Human Pose \citep{andriluka20142d} and Leeds Sport Poses (LSP) + extended training set \citep{Johnson10}. There are around 18k training images and 7k testing images involved in MPII. We split training set randomly to make a 3k size validation set while the rest are used for training. Results on LSP are reported by fine tuning the same model with the 11k extended LSP training set.

\textbf{CNN Training for DenseReg} 

We have used two different network architectures for our experiments. In particular, in order to be directly comparable to the DeepLab-v2 network in semantic segmentation experiments we first used a ResNet101~\citep{He2015} architecture with dilated convolutions ( atrous )~\citep{CP2015Semantic,mallat1999wavelet}, such that the stride of the CNN is $8$ and (b) an Hourglass-type network \cite{newell2016stacked}. We use bilinear interpolation to upscale both the $\hat{q}$ and $\hat{r}$ branches before the losses. The losses are applied at the input image scale and back-propagated through interpolation. We apply a weight to the smooth $L1$ loss layers to balance their contribution. In our experiments, we have used a weight of $40$ for quantized~($d=0.1$) and a weight of $70$ for non-quantized regression, which are determined by a coarse cross validation. 

For the ResNet based network, we use an initialization with a network pre-trained for the MS COCO segmentation task~\citep{lin2014microsoft}. The new layers are initialized with random weights drawn from Gaussian distributions. Large weights of the regression losses can be problematic at initialization even with moderate learning rates. To cope with this, we use initial training with a lower learning rate for a \textit{warm start} for a few iterations. We then use a base learning rate of $0.001$ with a polynomial decay policy for $20k$ iterations with a batch size of $10$ images.

For the hourglass architecture, we adopt \cite{newell2016stacked} with inception-v2 module (a Figure describing the network can be found in the Appendix). Each deconvolution layer involved is using a dilated convolution following by a $3\times 3$ convolution layer with stride 1 and same output channels as input channels.

\textbf{CNN Training for DenseReg Cascade} 

For the DenseReg cascade architecture (i.e., end-to-end trainable dense shape regression and articulated pose estimation by means of landmark localisation) we used a stack of two hourglasses. The first hourglass network is the one described above. The second hourglass network is regressing to a  heatmap representation of facial landmarks/body joints (68-channel heatmap for the landmark localisation experiments and 16-channel heatmap of the body pose estimation experiments). We apply $L2$ loss to the heatmap regression. Weights are applied to balance losses of both first and second hourglasses to have equal contribution. During training, we are randomly scaling with ratio between 0.75 and 1.25, randomly rotating with angle -30 to 30 degree, and randomly cropping images of size $321\times321$ to $256\times256$.

%%%%%%%%%%%%%%%%%%%%%%%%%%%%%%%%%%
%%%%% SEMANTIC SEGMENTATION %%%%%%
%%%%%%%%%%%%%%%%%%%%%%%%%%%%%%%%%%
\subsection{Semantic Segmentation}
\label{sec:exp_semantic_segmentation}
% \vspace{-0.35cm}

\begin{figure}[h]
\vspace{-0.25cm}
\centering
\includegraphics[width=\linewidth ]{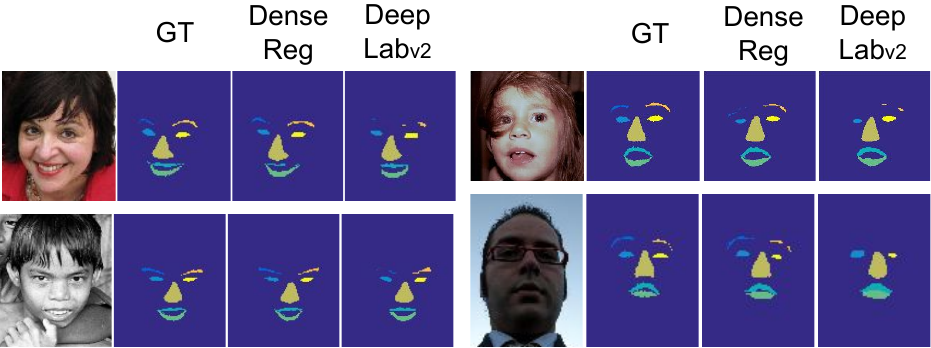}
\caption{Exemplar semantic segmentation results.}
\label{fig:Semantic}
\vspace{-0.25cm}

\end{figure}

As discussed in Sec.~\ref{sec:SDMs}, any labelling function defined on the template shape can be transferred to the image domain using the regressed coordinates. One application that can be naturally represented on the template shape is semantic segmentation of facial parts. 
To this end, we manually defined a segmentation mask of $8$ classes  (right/left eye, right/left eyebrow, upper/lower lip, nose, other) on the template shape, as shown in Fig.~\ref{fig:Semantic}. 
%%%%%%%%%%%%%%%%%%%%%%%%%%%%%%%%%%%%%%%%
%%%%%%%%%%%%%%%%%%%%%%%%%%%%%%%%%%%%%%%%

We compare against a state-of-the-art semantic part segmentation system (DeepLab-v2)~\citep{CP2016Deeplab} which is based on the same ResNet-101 architecture as our proposed DenseReg. We train DeepLab-v2 on the same training images (i.e. LFPW trainset, Helen trainset and AFW).  We generate the ground-truth segmentation labels for both training and testing images by transferring the segmentation mask using the ground-truth deformation-free coordinates explained in Sec.~\ref{sec:SDMs}. We employ the Helen testset~\citep{le2012interactive} for the evaluation.

Table~\ref{tab:Semantic} reports evaluation results using the intersection-over-union (IoU) ratio. Additionally, Fig.~\ref{fig:Semantic} shows some qualitative results for both methods, along with the ground-truth segmentation labels. The results indicate that the DenseReg outperforms DeepLab-v2. The reported improvement is substantial for several parts, such as eyebrows and lips. We believe that this result is significant given that DenseReg is not optimized for the specific task-at-hand, as opposed to DeepLab-v2 which was trained for semantic segmentation. This performance difference can be justified by the fact that DenseReg was exposed to a  richer label structure during training, which reflects the underlying variability and structure of the problem. 

%%%%%%%%%%%%%%%%%%%%%%%%%%%%%%%%%%%%%%%%

\begin{table}[h]
\centering
\begin{tabular}{|l|cc|}
\hline
\emph{Class} & \emph{Methods}& \\
\cline{2-3}
 & \textbf{DenseReg}  & Deeplab-v2 \\
\hline\hline
Left Eyebrow     & 48.35 & 40.57\\
Right Eyebrow    & 46.89 & 41.85\\
Left Eye         & 75.06 & 73.65\\
Right Eye        & 73.53 & 73.67\\
Upper Lip        & 69.52 & 62.04\\
Lower Lip        & 75.18 & 70.71\\
Nose             & 87.71 & 86.76\\
Other            & 99.44 & 99.37\\
\hline\hline
Average          & \textbf{71.96} & 68.58\\
\hline
\end{tabular}
\caption{Semantic segmentation accuracy on Helen testset measured using intersection-over-union (IoU) ratio.}
\vspace{-0.05cm}
\label{tab:Semantic}
\end{table}
%%%%%%%%%%%%%%%%%%%%%%%%%%%%%%%%%%%%%%%%

% We create a segmentation testing dataset by transferring the segmentation mask visualized in Fig.1 using the ground-truth deformation-free coordinates to the images.

% A standard way to approach this problem, given a dataset of segmented face parts, would be to use a fully convolutional neural network trained to classify each pixel. The alternative we propose is to transfer the labels to the image domain using the regressed coordinates. Using the segmentation ground-truth labels obtained, we train a state of the art semantic segmentation network based on ResNet-101\citep{Newdeeplab} architecture. Results from all three methods are visualized in Fig.~\ref{fig:Semantic} and the intersection-over-union measure for each class for both approaches are presented in Tab.\ref{tab:Semantic}. The results indicate that the dense regression framework does better on this task compared to the segmentation network. Even though our network was not optimized specifically for this task, it was exposed to a much richer label structure which explains the better performance.

%%%%%%%%%%%%%%%%%%%%%%%%%%%%%%%%%%
%%%%% LANDMARK LOCALIZATION %%%%%%
%%%%%%%%%%%%%%%%%%%%%%%%%%%%%%%%%%
\captionsetup[subfigure]{labelformat=empty}

\begin{figure*}[h]
\centering

\subfloat{\includegraphics[width=0.13775\textwidth]{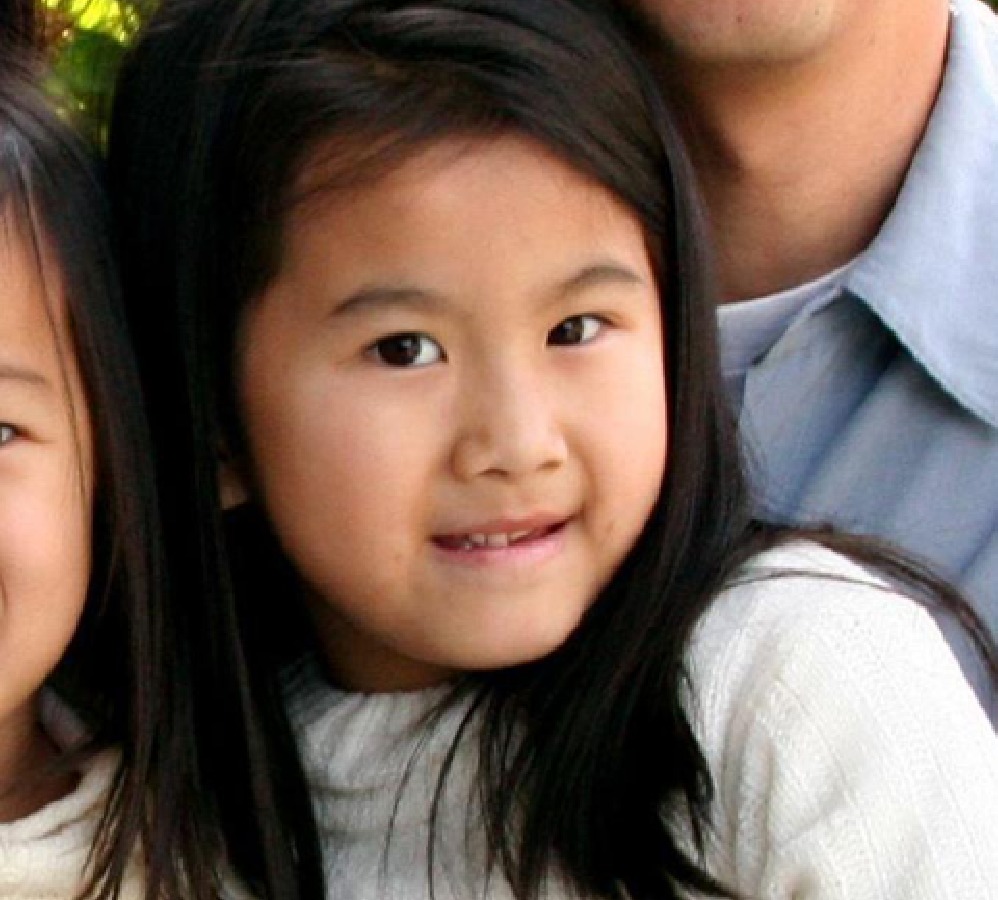}}\hspace{0.0005cm}
\subfloat{\includegraphics[width=0.13775\textwidth]{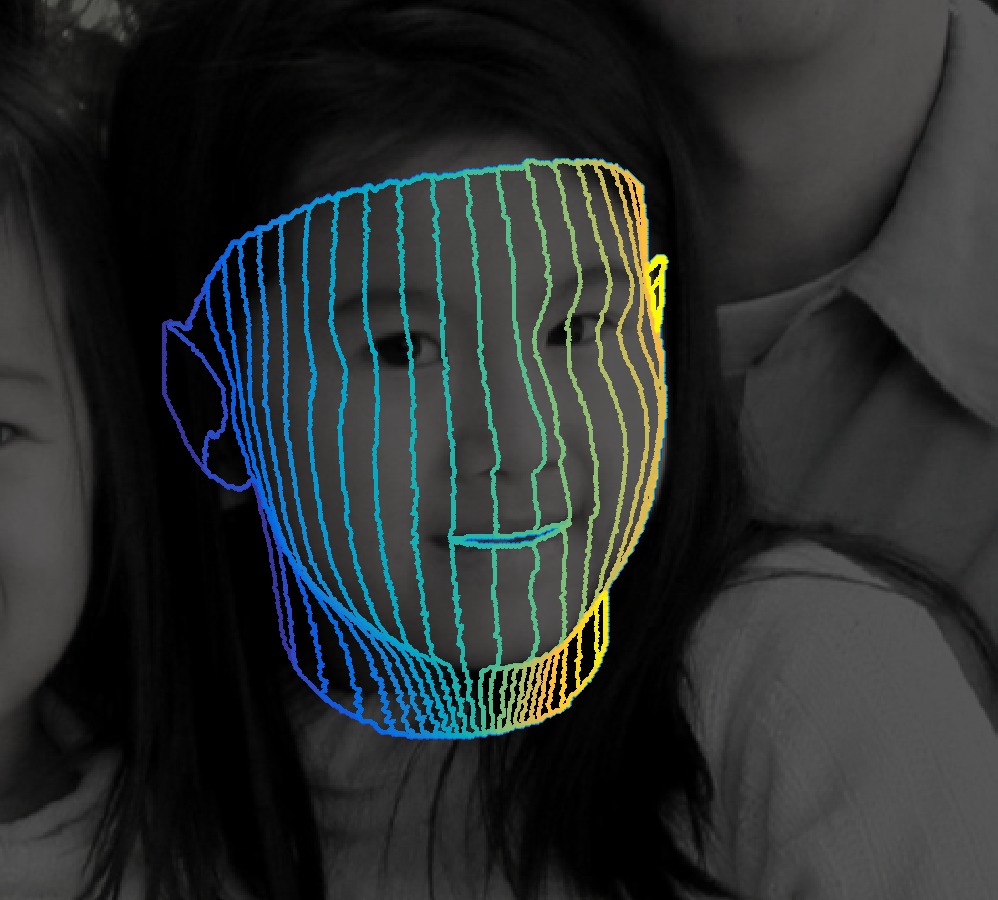}}\hspace{0.0005cm}
\subfloat{\includegraphics[width=0.13775\textwidth]{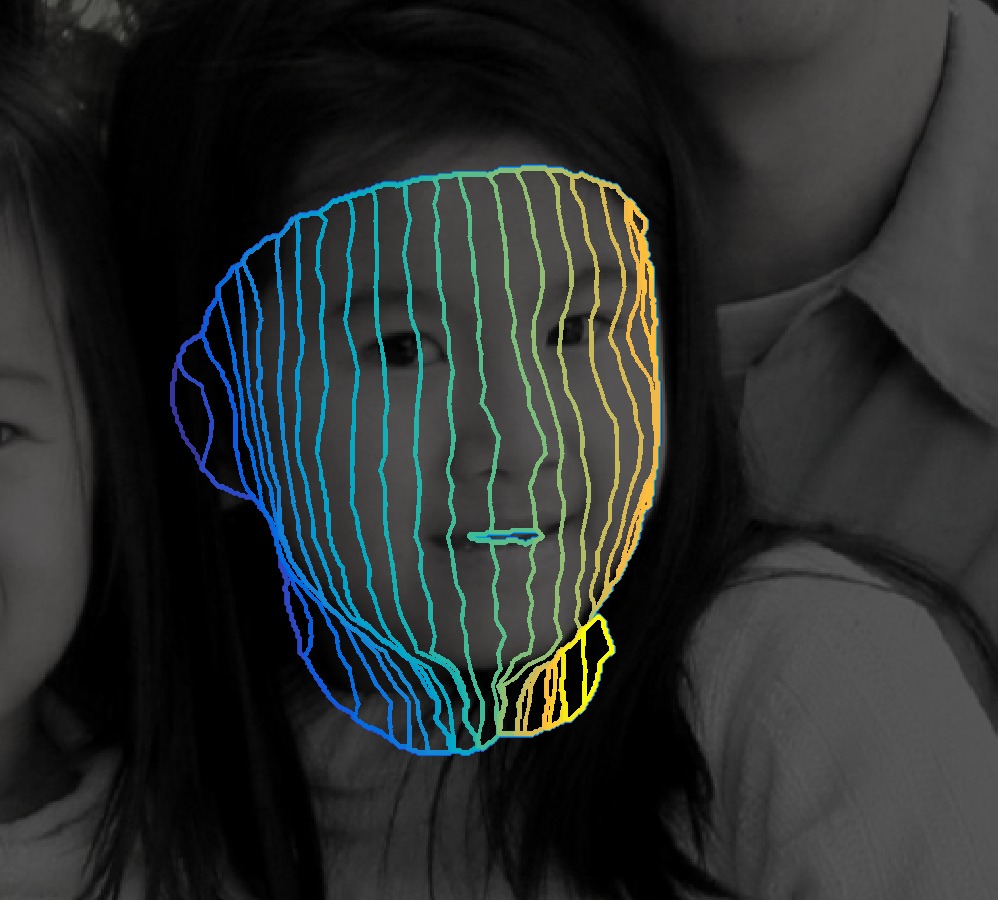}}\hspace{0.0005cm}
\subfloat{\includegraphics[width=0.13775\textwidth]{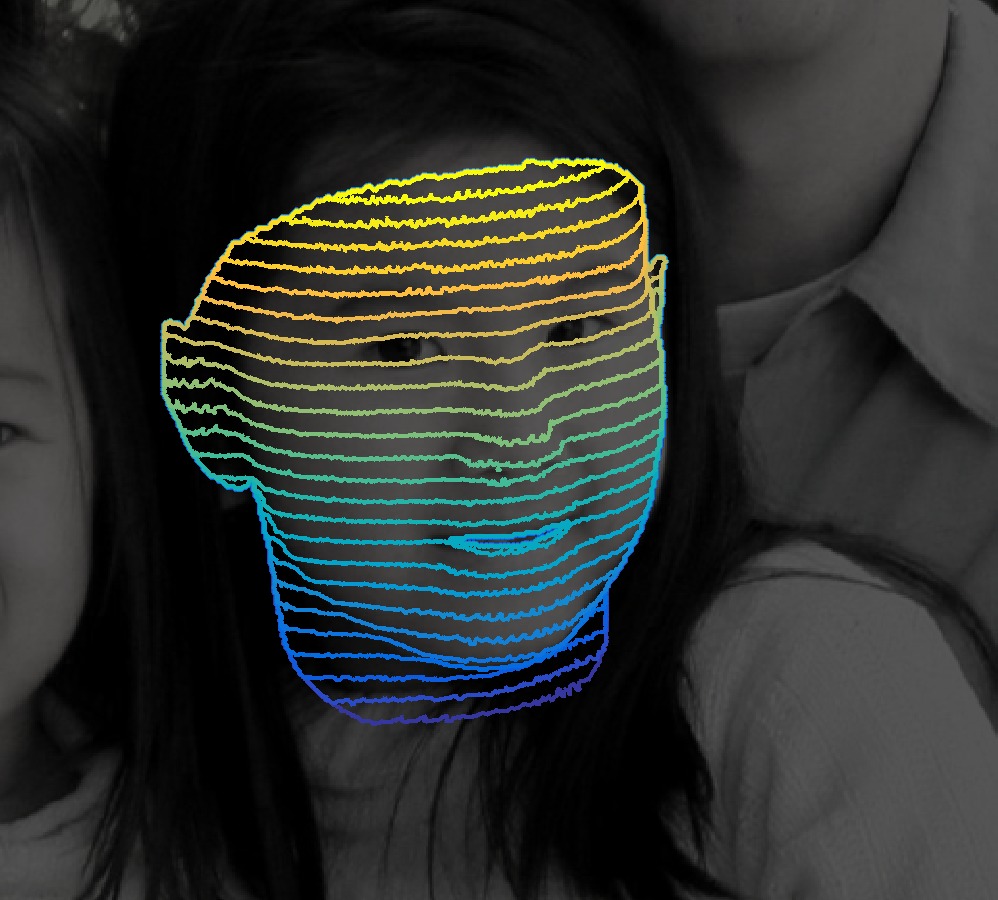}}\hspace{0.0005cm}
\subfloat{\includegraphics[width=0.13775\textwidth]{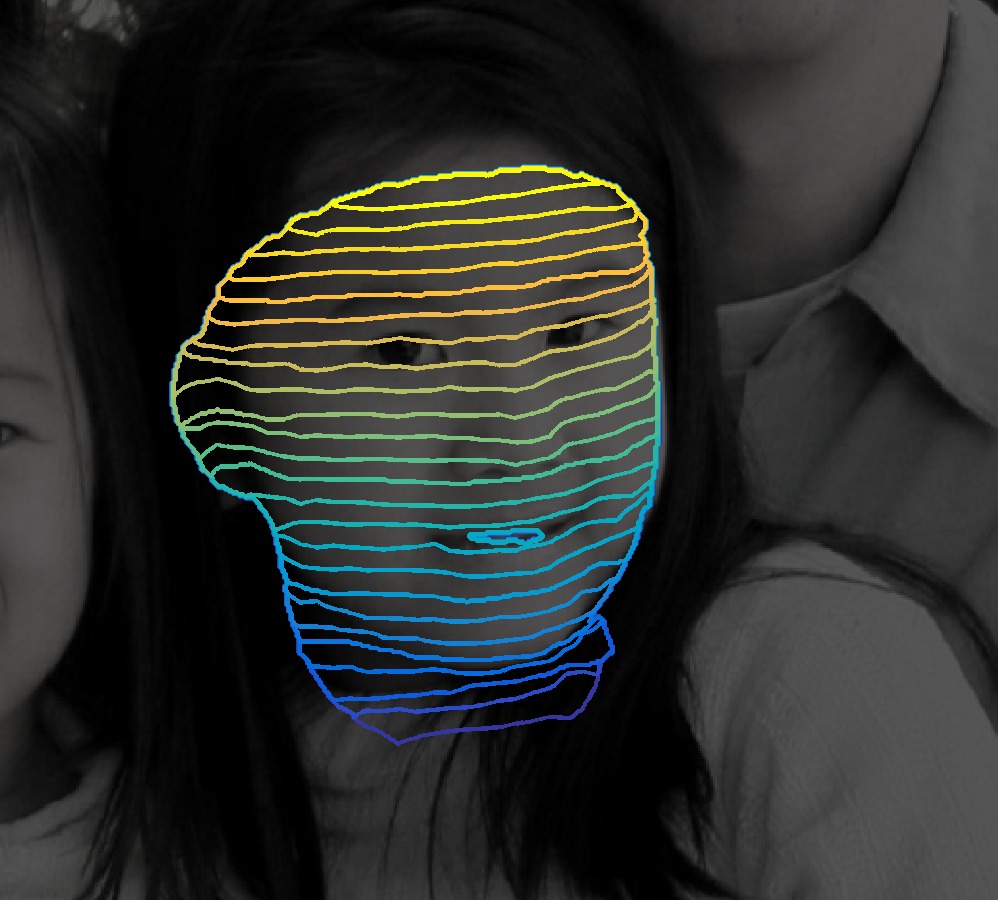}}\hspace{0.0005cm}
\subfloat{\includegraphics[width=0.13775\textwidth]{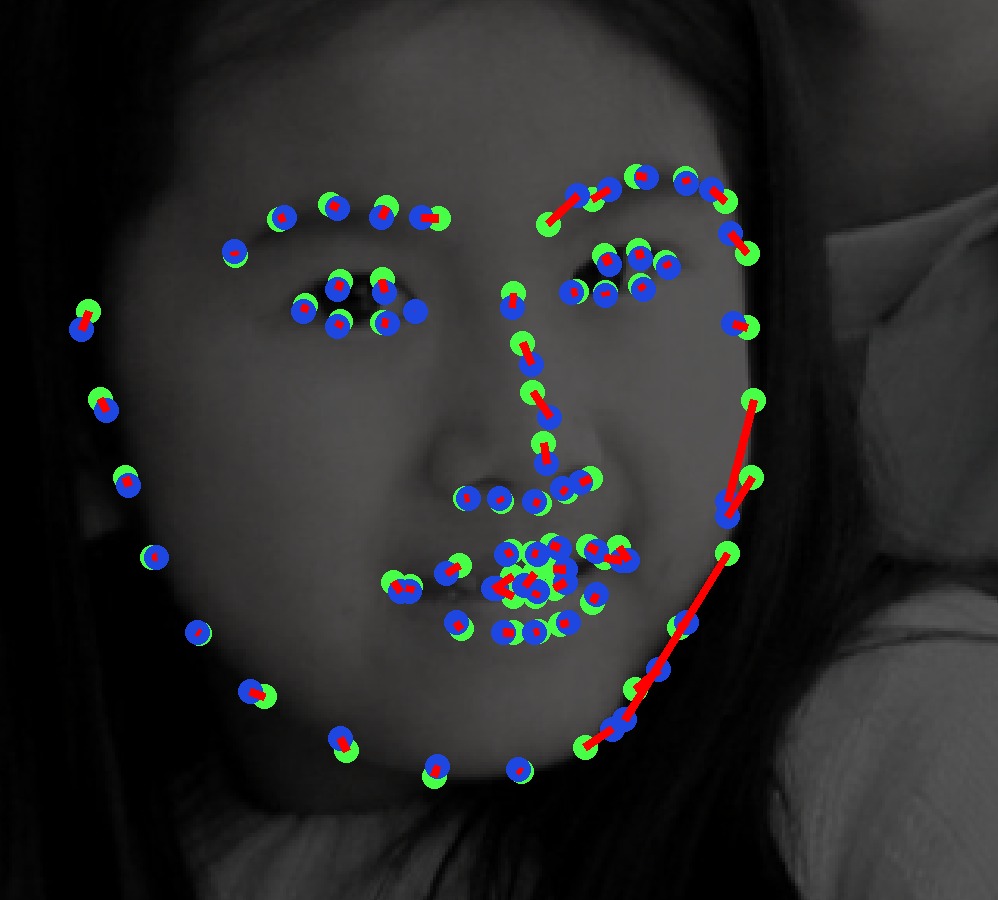}}\hspace{0.0005cm}
\subfloat{\includegraphics[width=0.13775\textwidth]{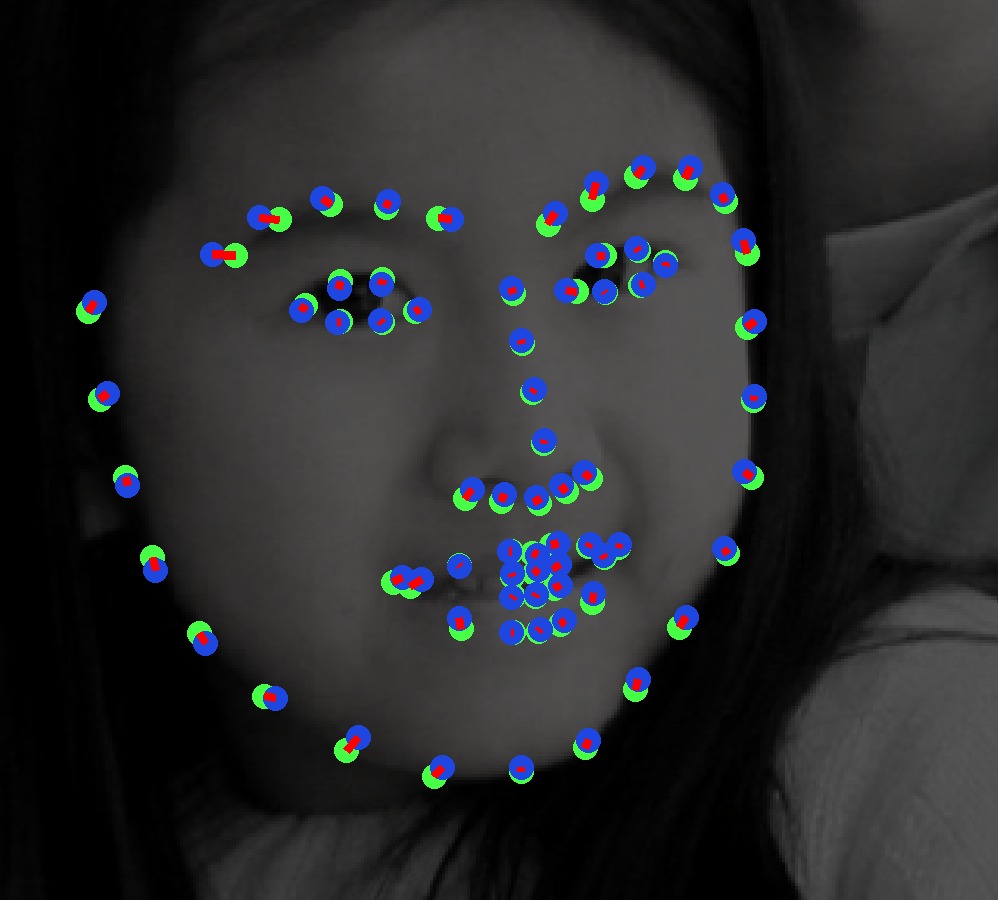}}\\
\vspace{-0.3cm}
% \vspace{-0.3cm}
\subfloat{\includegraphics[width=0.13775\textwidth]{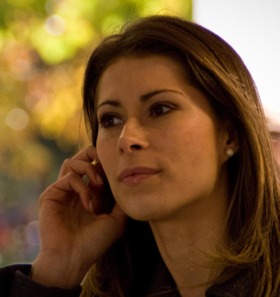}}\hspace{0.0005cm}
\subfloat{\includegraphics[width=0.13775\textwidth]{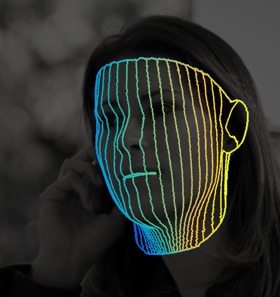}}\hspace{0.0005cm}
\subfloat{\includegraphics[width=0.13775\textwidth]{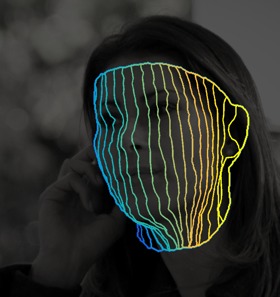}}\hspace{0.0005cm}
\subfloat{\includegraphics[width=0.13775\textwidth]{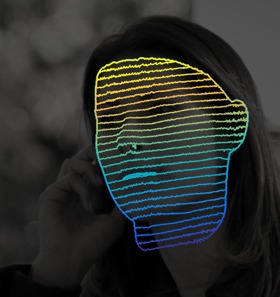}}\hspace{0.0005cm}
\subfloat{\includegraphics[width=0.13775\textwidth]{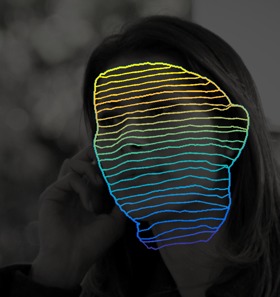}}\hspace{0.0005cm}
\subfloat{\includegraphics[width=0.13775\textwidth]{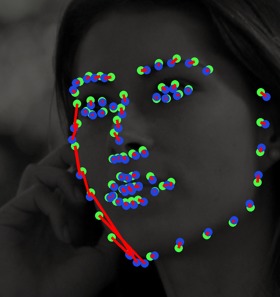}}\hspace{0.0005cm}
\subfloat{\includegraphics[width=0.13775\textwidth]{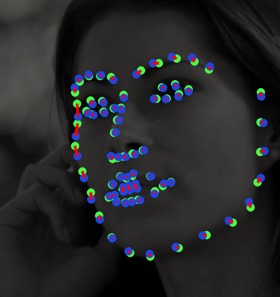}}\\
\vspace{-0.3cm}
\subfloat[Input Image]{\includegraphics[width=0.13775\textwidth]{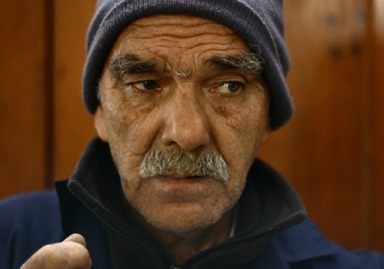}}\hspace{0.0005cm}
\subfloat[Groundtruth~$U$~$u^h$ ]{\includegraphics[width=0.13775\textwidth]{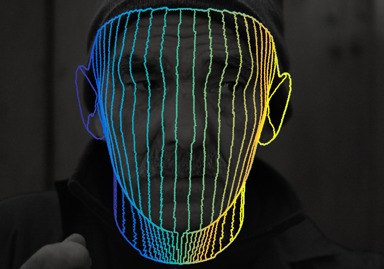}}\hspace{0.0005cm}
\subfloat[Estimated~$U$~$\hat{u}^h$ ]{\includegraphics[width=0.13775\textwidth]{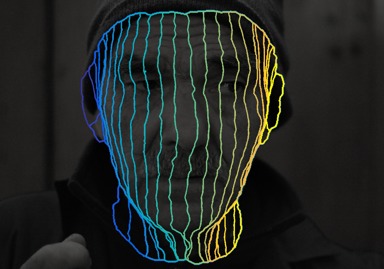}}\hspace{0.0005cm}
\subfloat[Groundtruth~$V$~$u^v$]{\includegraphics[width=0.13775\textwidth]{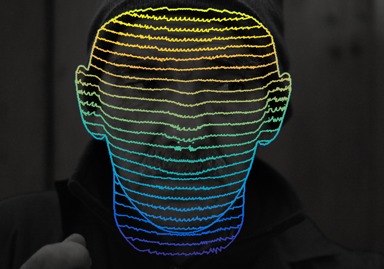}}\hspace{0.0005cm}
\subfloat[Estimated~$V$~$\hat{u}^v$]{\includegraphics[width=0.13775\textwidth]{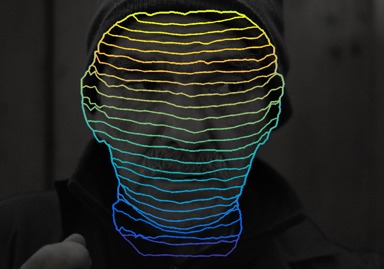}}\hspace{0.0005cm}
\subfloat[DenseReg]{\includegraphics[width=0.13775\textwidth]{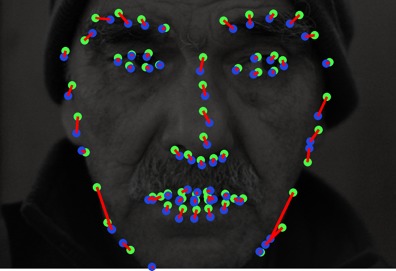}}\hspace{0.0005cm}
\subfloat[DenseReg+MDM]{\includegraphics[width=0.13775\textwidth]{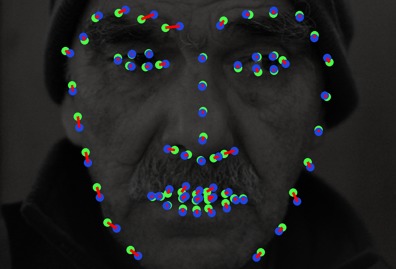}}\\
\vspace{-0.13cm}

% \subfloat{\includegraphics[width=0.13775\textwidth]{Figures/Results/8/I}}\hspace{0.0005cm}
% \subfloat{\includegraphics[width=0.13775\textwidth]{Figures/Results/8/HGT}}\hspace{0.0005cm}
% \subfloat{\includegraphics[width=0.13775\textwidth]{Figures/Results/8/Hreg}}\hspace{0.0005cm}
% \subfloat{\includegraphics[width=0.13775\textwidth]{Figures/Results/8/VGT}}\hspace{0.0005cm}
% \subfloat{\includegraphics[width=0.13775\textwidth]{Figures/Results/8/Vreg}}\hspace{0.0005cm}
% \subfloat{\includegraphics[width=0.13775\textwidth]{Figures/Results/8/Landmarks_alp}}\hspace{0.0005cm}
% \subfloat{\includegraphics[width=0.13775\textwidth]{Figures/Results/8/Landmarks_MDM}}\\

\caption{Qualitative Results. Ground-truth and estimated deformation-free coordinates and  landmarks obtained from DenseReg and DenseReg+MDM are presented. Estimated landmarks(blue), ground-truth(green), lines between estimated and ground-truth landmarks(red).}
\label{fig:qualitative}
\vspace{-0.3cm}

\end{figure*}

\subsection{Landmark Localization on Static Images}
\label{sec:exp_landmark_localization}

%%%%%%%%%%%%%%%%%%%%%%%%%%%%%%%%%%%%%%%%
\begin{figure}[!b]
\vspace{-0.25cm}
\centering
\includegraphics[width=0.9\linewidth]{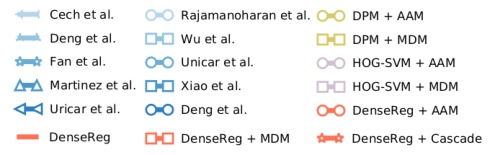}
\includegraphics[width=0.9\linewidth]{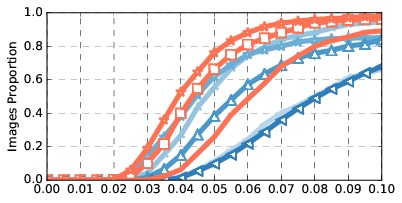}
\includegraphics[width=0.9\linewidth]{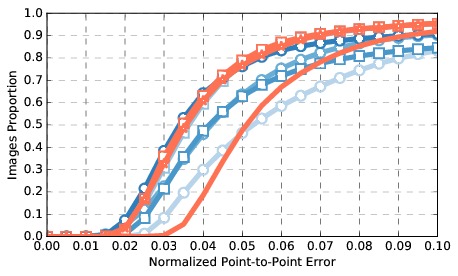}
\includegraphics[width=0.9\linewidth]{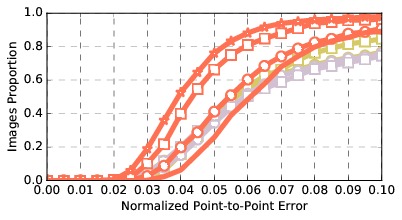}
\caption{Landmark localization results using 68 points. Accuracy is reported as Cumulative Error Distribution of RMS point-to-point error normalized with interocular distance. \emph{Top:} Comparison with state-of-the-art on the 300W testing dataset. \emph{Middle:} Comparison with state-of-the-art on the 300WV tracking dataset.\emph{Bottom:} Self-evaluation results on the 300W testing dataset.}
\vspace{-0.15cm}
\label{fig:300w}
\end{figure}
%%%%%%%%%%%%%%%%%%%%%%%%%%%%%%%%%%%%%%%%

DenseReg can be readily used for the task of facial landmark localization on static images. Given the landmarks' locations on the template shape, it is straightforward to estimate the closest points in the deformation-free coordinates on the images. The local minima of the Euclidean distance between the estimated coordinates and the landmark coordinates are considered as detected landmarks. In order to find the local minima, we simply analyze the connected components separately. Even though more sophisticated methods for covering ``touching shapes'' can be used, we found that this simplistic approach is sufficient for the task. 

Note that the closest deformation-free coordinates among all \emph{visible} pixels to a landmark point is not necessarily the correct corresponding landmark. This phenomenon is called ``landmark marching''~\citep{zhu2015high} and mostly affects the jaw landmarks which are dependent on changes in head pose. It should be noted that we do not use any explicit supervision for landmark detection nor focus on ad-hoc methods to cope with this issue. Errors on jaw landmarks due to invisible coordinates and improvements thanks to deformable models can be observed in Fig.~\ref{fig:qualitative}.

Herein, we evaluate the landmark localization performance of DenseReg as well as the performance obtained by employing DenseReg as an initialization for deformable models~\citep{papandreou2008adaptive,tzimiropoulos2014gauss,antonakos2015feature,trigeorgis2016mnemonic} trained for the specific task. In the second scenario, we provide a slightly improved initialization with very small computational cost by reconstructing the detected landmarks with a PCA shape model that is constructed from ground-truth annotations.

We present experimental results using the very challenging 300W benchmark. This is the testing database that was used in the 300W competition~\citep{sagonas_iccv_300w_2013,sagonas2016300} - the most important facial landmark localization challenge. The error is measured using the point-to-point RMS error normalized with the interocular distance and reported in the form of Cumulative Error Distribution (CED). Figure~\ref{fig:300w} (bottom) presents some self-evaluations in which we compare the quality of initialization for deformable modelling between DenseReg \footnote{We have tested both the ResNet and the hourglass-based architecture and the provided the same results, hence we will not present them separately.} and two other standard face detection techniques (HOG-SVM~\citep{king2015max}, DPM~\citep{mathias2014face}). The employed deformable models are the popular generative approach of patch-based Active Appearance Models (AAM)~\citep{papandreou2008adaptive,tzimiropoulos2014gauss,antonakos2015feature}, as well as the current state-of-the-art approach of Mnemonic Descent Method (MDM)~\citep{trigeorgis2016mnemonic}. It is interesting to notice that the performance of DenseReg without any additional deformable model on top, already outperforms even HOG-SVM detection combined with MDM. Especially when DenseReg is combined with MDM, it greatly outperforms all other combinations.

Figure~\ref{fig:300w} (top) compares DenseReg + cascade, DenseReg + MDM and DenseReg with the results of the latest 300W competition~\citep{sagonas2016300}. 
We greatly outperform all competitors by a large margin. It should be noted that the participants of the competition did not have any restrictions on the amount of training data employed and some of them are industrial companies (e.g. Fan etal.~\citep{fan2016approaching}), which further illustrates the effectiveness of our approach. Finally, Table~\ref{tab:300w} reports the area under the curve (AUC) of the CED curves, as well as the failure rate for a maximum RMS error of $0.1$. 

Even though both DenseReg and MDM are based on convolutional architectures DenseReg plus MDM is not a fully end-to-end trainable architecture. On the other hand the proposed DenseReg cascade architecture is a end-to-end trainable. It is evident Table \ref{tab:300w} that DenseReg cascade largly outperfoms all other tested methods achieving a new performance record for 300W test set. Apart from the accuracy improvement shown by the AUC, we believe that the reported failure rate of $2.67\%$ is remarkable and highlights the robustness of DenseReg.

%%%%%%%%%%%%%%%%%%%%%%%%%%%%%%%%%%%%%%%%
\begin{table}[h]
\vspace{-0.05cm}

\centering
\scalebox{0.9}{
\begin{tabular}{|l|c|c|}
\hline
\emph{Method} & \emph{AUC} & \emph{Failure Rate (\%)}\\
\hline\hline
\textbf{DenseReg Cascade}             & \textbf{0.5702} & \textbf{2.17} \\
\textbf{DenseReg + MDM}                 & \textbf{0.5219} & \textbf{3.67} \\
DenseReg                                & 0.3605 & 10.83 \\
\cite{fan2016approaching}    & 0.4802 & 14.83 \\
\cite{deng2016m}            & 0.4752 & 5.5   \\
\cite{martinez20162}    & 0.3779 & 16.0  \\
\cite{vcech2016view}       & 0.2218 & 33.83 \\
\cite{uvrivcavr2016view} & 0.2109 & 32.17 \\
\hline
\end{tabular}
}
\caption{Summary of landmark localization results on the 300W testing dataset using 68 points. Accuracy is reported as the AUC and the Failure Rate.}
\label{tab:300w}
\vspace{-0.15cm}
\end{table}
%%%%%%%%%%%%%%%%%%%%%%%%%%%%%%%%%%%%%%%%

%%%%%%%%%%%%%%%%%%%%%%%%%%%%%%%%%%
%%%%%% DEFORMABLE TRACKING %%%%%%%
%%%%%%%%%%%%%%%%%%%%%%%%%%%%%%%%%%
\subsection{Deformable Tracking}
\label{sec:exp_deformable_tracking}

%%%%%%%%%%%%%%%%%%%%%%%%%%%%%%%%%%%%%%%%
\begin{table*}[h]
\centering
\scalebox{0.9}{
\begin{tabular}{|l|c|c|}
\hline
\emph{Method} & \emph{AUC} & \emph{Failure Rate (\%)}\\
\hline\hline
\textbf{DenseReg Cascade}                        & \textbf{0.5853} & \textbf{4.36} \\
\textbf{DenseReg + MDM}                            & \textbf{0.5937} & \textbf{4.57} \\
DenseReg                                           & 0.4320 & 8.1   \\
\cite{yang2015facial}                  & 0.5832 & 4.66  \\
\cite{xiao2015facial}                  & 0.5800 & 9.1   \\
\cite{rajamanoharan2015view} & 0.5154 & 9.68  \\
\cite{wu2015shape}                       & 0.4887 & 15.39 \\
\cite{uricar2015facial}              & 0.4059 & 16.7  \\
\hline
\end{tabular}
 }
\caption{Deformable tracking results against the state-of-the-art on the 300VW testing dataset using 68 points. Accuracy is reported as AUC and the Failure Rate.}
\vspace{-0.25cm}
\label{tab:300vw}
\end{table*}
%%%%%%%%%%%%%%%%%%%%%%%%%%%%%%%%%%%%%%%%

For the challenging task of deformable face tracking on lengthy videos, we employ the testing database of the 300VW challenge~\citep{300VW,chrysos2015offline} - the only existing benchmark for deformable tracking ``in-the-wild''. The benchmark consists of $114$ videos ($\sim 218k$ frames in total) and includes videos captured in totally arbitrary conditions (severe occlusions and extreme illuminations).
%that are separated into three categories: \emph{(i)}~Videos captured in well-lit environments without occlusions, \emph{(ii)}~videos captured in unconstrained illumination conditions, and \emph{(iii)}~videos captured in totally arbitrary conditions (severe occlusions and extreme illuminations).  
%Similar to the landmark localization case, performance is reported as the CED of the RMS point-to-point errors normalized with the interocular distance. 
The tracking is performed based on sparse landmark points, thus we follow the same strategy as in the case of landmark localization in Sec.~\ref{sec:exp_landmark_localization}.

We compare the output of DenseReg, as well as DenseReg + MDM and DenseReg + Cascade which was the best performing combination for landmark localization in static images (Sec.~\ref{sec:exp_landmark_localization}), against the participants of the 300VW challenge.

Table~\ref{tab:300vw} reports the AUC and Failure Rate measures. DenseReg + Cascade and DenseReg + MDM demonstrates better performance than the winner of the 300VW competition. It should be highlighted that our approach is not fine-tuned for the task-at-hand as opposed to the rest of the methods that were trained on video sequences and most of them make some kind of temporal modelling. Finally, similar to the 300W case, the participants were allowed to use unlimited training data (apart from the provided training sequences), as opposed to DenseReg (Cascade and MDM) that were trained only on the $3148$ images mentioned in Sec.~\ref{sec:training_setup}. DenseReg cascade architecture has achieved similar performance to DenseReg + MDM (this is mainly because the results in challenge one are currently saturated), as measured in AUC, but achieved a lower failure rate.

%%%%%%%%%%%%%%%%%%%%%%%%%%%
%%%%%% Pose TRACKING %%%%%%
%%%%%%%%%%%%%%%%%%%%%%%%%%%
\subsection{Dense Correspondence for the Human Body}
\label{sec:exp_human}

\begin{figure*}[t!]
    \centering
    \newcommand{\flowh}{0.28\columnwidth}
    \includegraphics[height=\flowh]{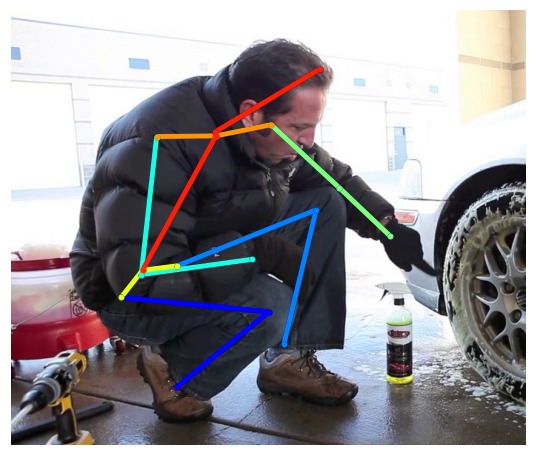}
    \hfill
    \includegraphics[height=\flowh]{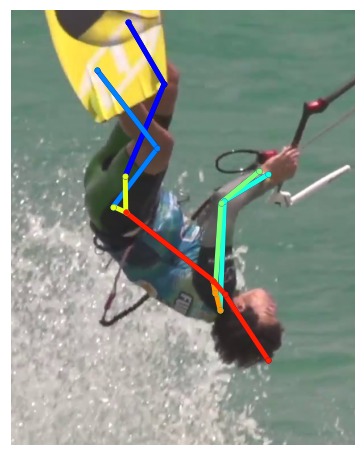}
    \hfill
    \includegraphics[height=\flowh]{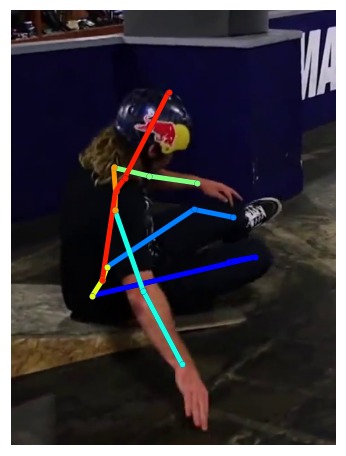}
    \hfill
    \includegraphics[height=\flowh]{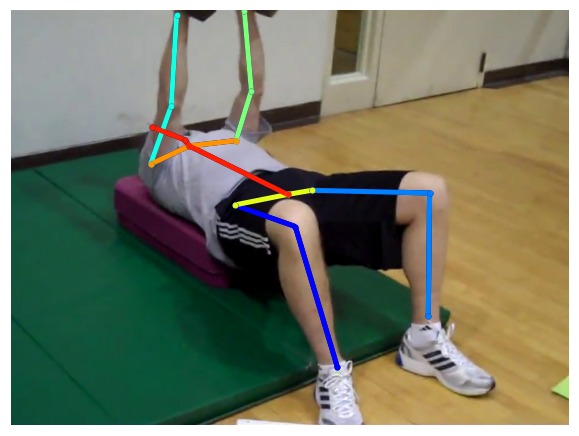}
    \hfill
    \includegraphics[height=\flowh]{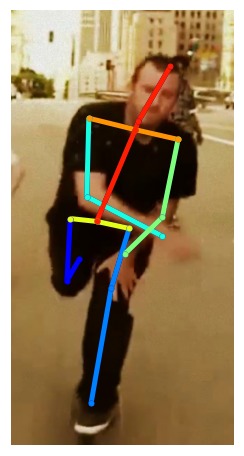}
    \hfill
    \includegraphics[height=\flowh]{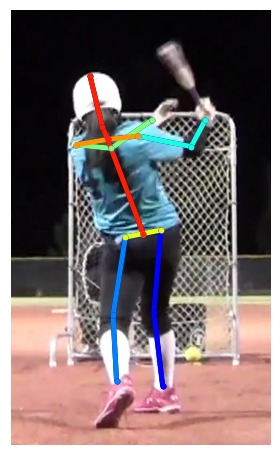}
    \hfill
    \includegraphics[height=\flowh]{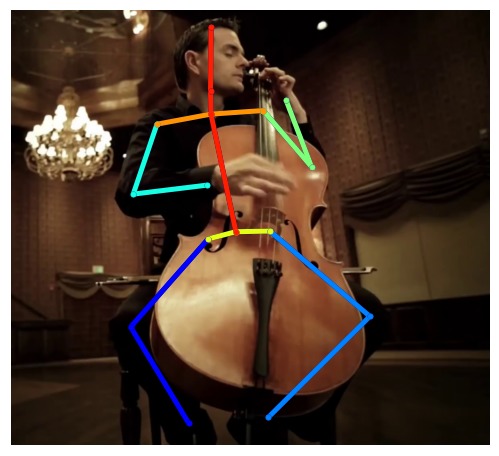}
    \\
    \includegraphics[height=\flowh]{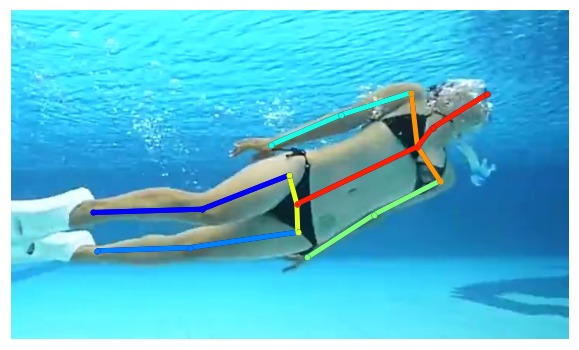}
    \hfill
    \includegraphics[height=\flowh]{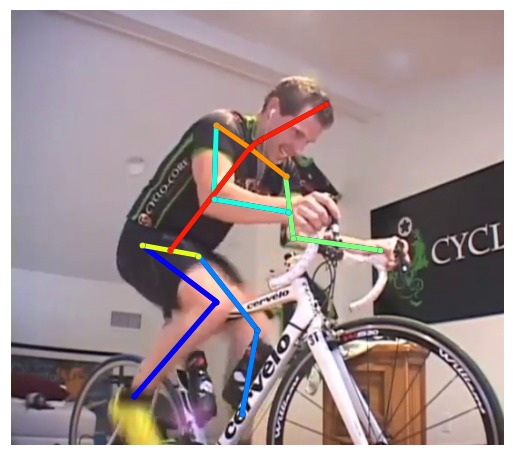}
    \hfill
    \includegraphics[height=\flowh]{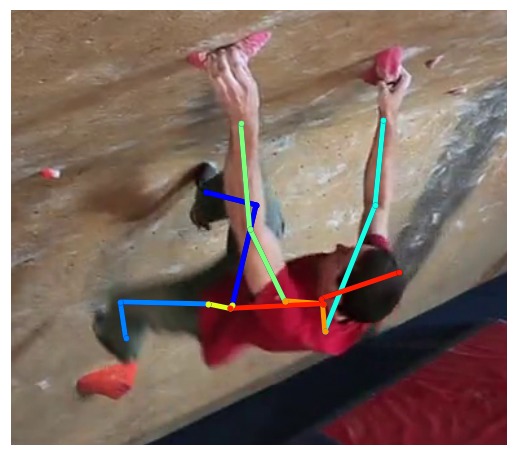}
    \hfill
    \includegraphics[height=\flowh]{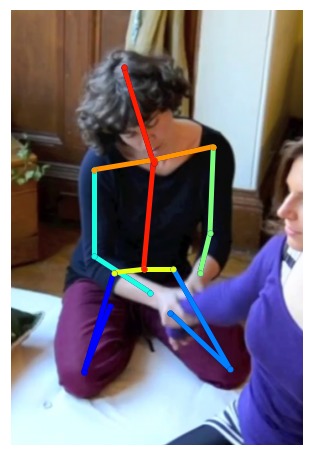}
    \hfill
    \includegraphics[height=\flowh]{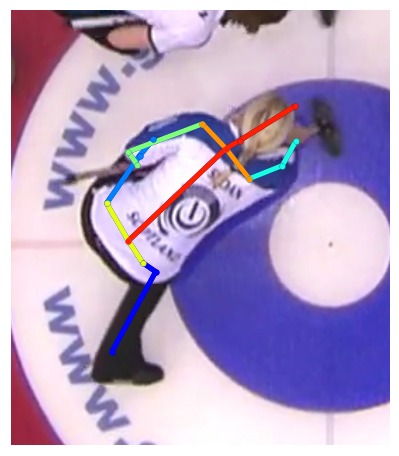}
    \hfill
    \includegraphics[height=\flowh]{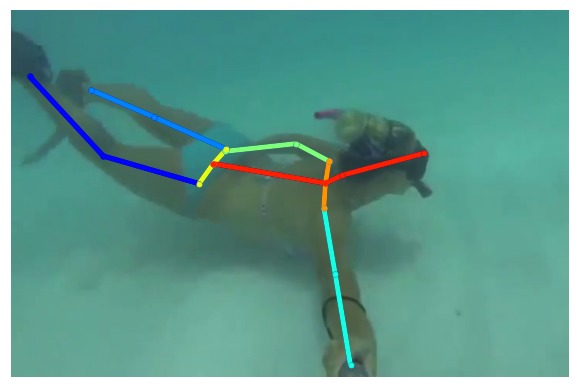}
    \\
    \newcommand{\flowhh}{0.27\columnwidth}
    \includegraphics[height=\flowhh]{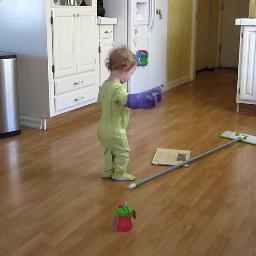}
    \hfill
    \includegraphics[height=\flowhh]{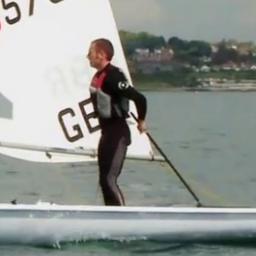}
    \hfill
    \includegraphics[height=\flowhh]{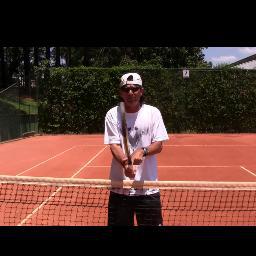}
    \hfill
    \includegraphics[height=\flowhh]{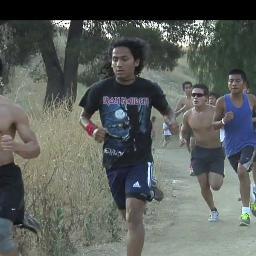}
    % \hfill
    % \includegraphics[height=\flowhh]{Figures/pose/qualitative/view/view_img_101}
    \hfill
    \includegraphics[height=\flowhh]{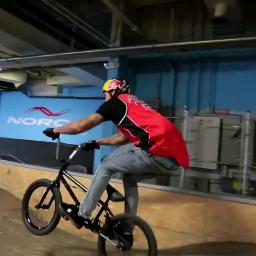}
    \hfill
    \includegraphics[height=\flowhh]{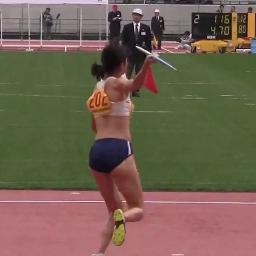}
    % \hfill
    % \includegraphics[height=\flowhh]{Figures/pose/qualitative/view/view_img_121}
    \hfill
    \includegraphics[height=\flowhh]{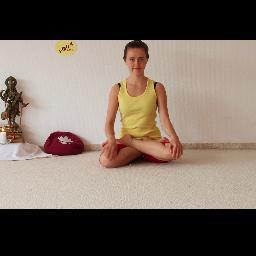}
    \\
    \includegraphics[height=\flowhh]{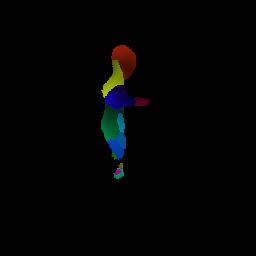}
    \hfill
    \includegraphics[height=\flowhh]{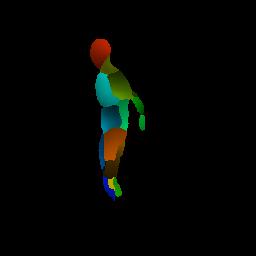}
    \hfill
    \includegraphics[height=\flowhh]{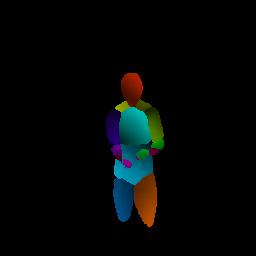}
    \hfill
    \includegraphics[height=\flowhh]{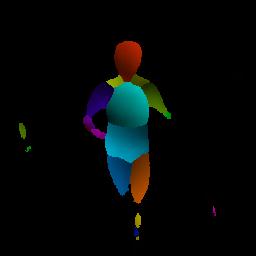}
    % \hfill
    % \includegraphics[height=\flowhh]{Figures/pose/qualitative/view/view_iuv_101}
    \hfill
    \includegraphics[height=\flowhh]{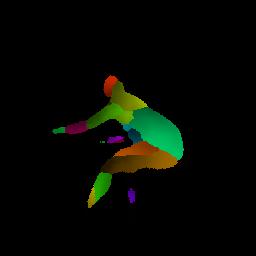}
    \hfill
    \includegraphics[height=\flowhh]{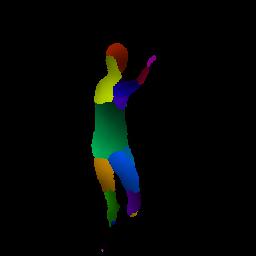}
    % \hfill
    % \includegraphics[height=\flowhh]{Figures/pose/qualitative/view/view_iuv_121}
    \hfill
    \includegraphics[height=\flowhh]{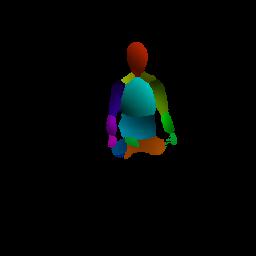}
    
    \caption{Examplar joints localisations on MPII and LSP test set. \textit{TOP 2 Rows}: Predictions of challenging poses in MPII and LSP test set. \textit{Bottom 3 row}: Examplar predictions of Landmarks, together with estimated dense correspondence IUV outputs are shown at each row correspondingly. Figure best viewed by zooming in.}
    \label{fig:qualitative_pose}
    \vspace{12pt}
\end{figure*}
\begin{table*}[t!]
\small
\begin{center}
\begin{tabular}{|l|r|r|r|r|r|r|r|r|r|}
\hline
\textbf{Methods} &\textbf{Head}   & \textbf{Shoulder} & \textbf{Elbow} & \textbf{Wrist} & \textbf{Hip}   & \textbf{Knee} & \textbf{Ankle} & \textbf{Total} & \textbf{AUC} \\
\hline
\hline
% Chu et al., CVPR'17           & 98.5  & 96.3  & 91.9  & 88.1  & 90.6  & 88.0 & 85.0 & 91.5 & 63.8 \\
\cite{newell2016stacked}       & 98.2  & 96.3  & 91.2  & 87.1  & \textbf{90.1}  & 87.4 & 83.6 & 90.9 & \textbf{62.9} \\
 \cite{bulat2016human} & 97.9  & 95.1  & 89.9  & 85.3  & 89.4  & 85.7 & 81.7 & 89.7 & 59.6 \\
 \cite{wei2016convolutional}          & 97.8  & 95.0  & 88.7  & 84.0  & 88.4  & 82.8 & 79.4 & 88.5 & 61.4 \\
 \cite{pishchulin16cvpr}   & 94.1  & 90.2  & 83.4  & 77.3  & 82.6  & 75.7 & 68.6 & 82.4 & 56.5 \\
\hline
DenseReg Cascade & \textbf{98.5}  & \textbf{96.4}  & \textbf{92.1}  & \textbf{88.2}  & 89.4  & \textbf{88.6} & \textbf{85.6} & \textbf{91.6} & 62.8 \\
\hline
\end{tabular}
\end{center}
\caption{Joints Localisation Accuracy on MPII dataset.}
\label{tab:mpii}
\vspace{12pt}
\end{table*}
\begin{table*}[t!]
\small
\begin{center}
\begin{tabular}{|l|r|r|r|r|r|r|r|r|r|}
\hline
\textbf{Methods} &\textbf{Head}   & \textbf{Shoulder} & \textbf{Elbow} & \textbf{Wrist} & \textbf{Hip}   & \textbf{Knee} & \textbf{Ankle} & \textbf{Total} & \textbf{AUC} \\
\hline\hline
 \cite{pishchulin2013poselet} & 87.2  & 56.7  & 46.7  & 38.0  & 61.0  & 57.5 & 52.7 & 57.1 & 35.8\\
 \cite{wei2016convolutional} & \textbf{97.8}  & 92.5  & 87.0  & 83.9  & 91.5  & 90.8 & 89.9 & 90.5 & 65.4\\
 \cite{bulat2016human}& 97.2  & 92.1  & 88.1  & 85.2  & \textbf{92.2}  & 91.4 & 88.7 & 90.7 & 63.4\\
 \cite{insafutdinov16ariv}& 97.4  & 92.7  & 87.5  & 84.4  & 91.5  & 89.9 & 87.2 & 90.1 & \textbf{66.1}\\
\hline
%HG+IUV &93.7  & 92.0  & 87.2  & 85.2  & 90.5  & 91.1 & 89.3 & 89.9 & 63.0\\
DenseReg Cascade& 94.7  & \textbf{93.4}  & \textbf{90.2}  & \textbf{88.2}  & 91.8  & \textbf{92.9} & \textbf{91.6} & \textbf{91.8} & 65.6\\
\hline
\end{tabular}
\end{center}
\caption{Joints Localisation Accuracy on LSP dataset.}
\label{tab:lsp}
\vspace{12pt}
\end{table*}

\begin{figure}[!h]
\centering
\includegraphics[width=0.89\linewidth]{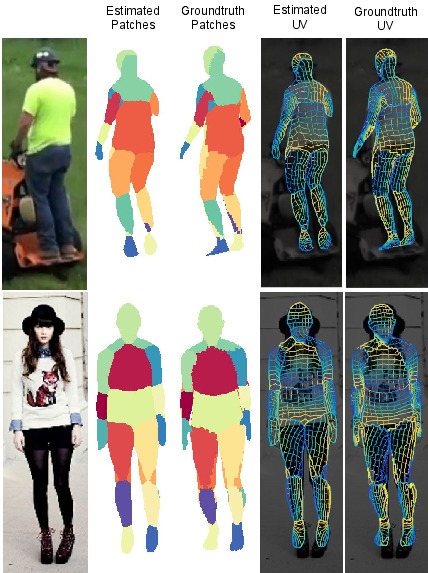}
\caption{Dense Correspondence for human body.}
\vspace{-0.25cm}
\label{fig:DenseReg_Human}
\end{figure}

Since there are no dense correspondence results between a 3D human model and image pixels in literature, we demonstrate the performance of our system through visual results from our test-set partition of the UP dataset in Fig.\ref{fig:DenseReg_Human}. In order to provide quantitative experimental results in the following we evaluate DenseReg cascade architecture for the problem of articulated body pose estimation. 

\noindent\textbf{Evaluation Metrics} The accuracies reported follow the Percentage Correct Keypoints (PCK) measurement on LSP dataset. Normalised PCK measurement by the scale of head (PCKh) is used for MPII on both validation and test set. Note that the performance gap between validation and test set is due to the use of invisible parts in measuring the performance. That is, in the validation set we measured the performance making use of the invisible parts, while the test set protocol of MPII does not use the invisible parts when computing the evaluation metrics.

\noindent\textbf{Model Training} Our model is implemented using TensorFlow~\footnote{https://tensorflow.org}. 15k images from the training set mentioned above are used with augmentations. Each pose instance in the image was cropped to size $384 \times 384$. Cropped images are then randomly flipped, rotated by $\pm 30^{\degree} $ and rescaled by $0.75$ to $1.25$ before cropping to size $256 \times 256$. The model are trained with initial learning rate $1\times10^{-3}$ with exponential decay factor of $0.97$ at every $2$ epochs. The models were trained for 100 epochs before testing.

\label{sec:pred_mpii_lsp}
\subsubsection{Joints Localisation on MPII \& LSP}

Results reported on MPII are obtained by using the proposed DenseReg cascade architecture. Table \ref{tab:mpii} provides a comparison of the proposed method with the state-of-the-art in MPII, while Table \ref{tab:lsp} provides a comparison with the state-of-the-art on LSP database.

Some qualitative results are collected in Fig~\ref{fig:qualitative} for the test sets of MPII and LSP. Top three rows show joint localization in challenging poses e.g. extreme viewing angles, challenging poses, occlusions, self occlusions and ambiguities. The bottom row demonstrates dense shape correspondence estimated on test images using the DenseReg cascade architecture.

%%%%%%%%%%%%%%%%%%%%%%%%%%%
%%%%%% EAR TRACKING %%%%%%%
%%%%%%%%%%%%%%%%%%%%%%%%%%%
\subsection{Ear Landmark Localization}
\label{sec:exp_ear}

\begin{figure}[h]
\centering
\includegraphics[width=\linewidth]{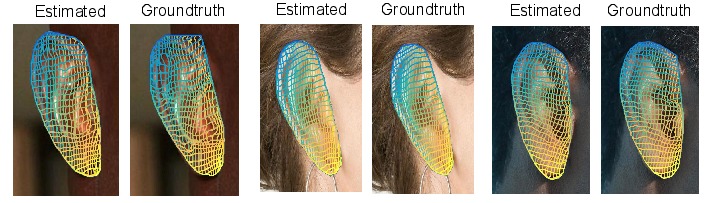}
\caption{Exemplar pairs of deformation-free coordinates of dense landmarks on human ear.}
\label{fig:ears_examples}
\end{figure}

We have also performed experiments on the human ear. We employ the $602$ images and sparse landmark annotations that were generated in a semi-supervised manner by Zhou et al.~\citep{Zhou_2016_CVPR}. Due to the lack of a 3D model of the human ear, we apply Thin Plate Splines to bring the images into dense correspondence and obtain the deformation-free space. We perform landmark localization following the same procedure as in Sec.~\ref{sec:exp_landmark_localization}.
%We split the images in $500$ for training and $102$ for testing.  
Quantitative results are detailed in the supplementary material, where we compare DenseReg, DenseReg + AAM and DenseReg + MDM with alternative DPM detector based initializations.  We observe that DenseReg results are highly accurate and clearly outperforms the DPM based alternative even without a deformable model. Examples for dense human ear correspondence estimated by our system  are presented in Fig.~\ref{fig:ears_examples}.

\subsubsection{Ear Shape Regression}

\begin{figure}[h]
\centering
  \includegraphics[trim={3.1cm 8cm 0.35cm 7.3cm},width=0.7\linewidth]{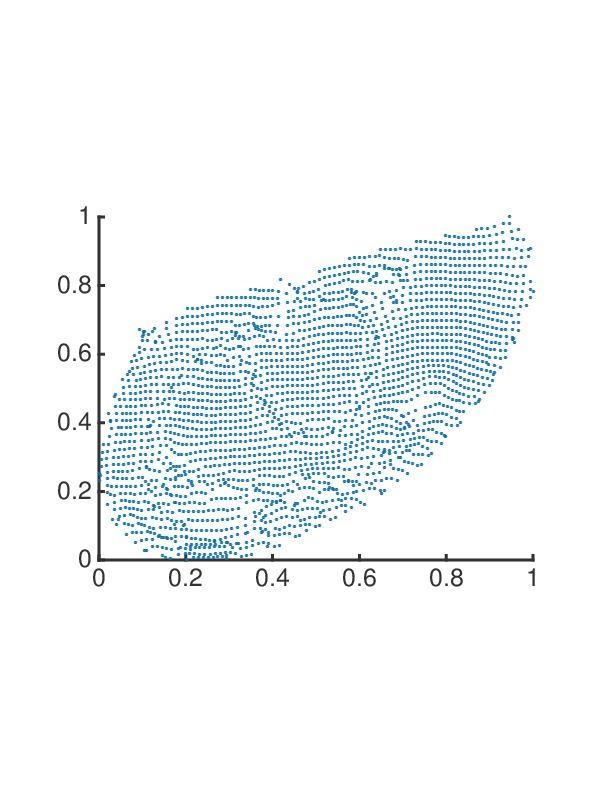}
\caption{ Deformation-free space for the template ear shape. }
\label{fig:earpoints}
\end{figure}

The deformation-free space for the ear shape template is visualized in Fig.~\ref{fig:earpoints}. The colouring of the qualitative results that are presented in the paper and this supplementary materials document are generated using these coordinates. On Table.\ref{tab:ears}, we provide failure rates and the Area Under Curve(AUC) measures based on the CED curve of the human ear landmark localization experiment, which were not provided in the paper due to space constraints. Further qualitative examples for regressed and ground-truth deformation-free ear coordinates are provided in Fig.~\ref{fig:ears_examples}.

%%%%%%%%%%%%%%%%%%%%%%%%%%%%%%%%%%%%%%%%
% \FloatBarrier
\begin{table}[h!]
\centering
\begin{tabular}{|l|c|c|}
\hline
\emph{Method} & \emph{AUC} & \emph{Failure Rate (\%)}\\
\hline\hline
\textbf{DenseReg + MDM} & \textbf{0.4842} & \textbf{0.98} \\
DenseReg       &  0.4150 &   1.96 \\
DenseReg + AAM &  0.4263 &  0.98 \\
DPM + MDM      &  0.4160 &  15.69 \\
DPM + AAM      &  0.3283 &  22.55 \\
\hline
\end{tabular}
\caption{Landmark localization results on human ear using 55 points. Accuracy is reported as the Area Under the Curve (AUC) and the Failure Rate of the Cumulative Error Distribution of the normalized RMS point-to-point error.}
\label{tab:ears}
\end{table}

\section{Conclusion}\label{S:CONCLUSIONS}
We propose a fully-convolutional regression approach for establishing dense correspondence fields between objects in natural images and three-dimensional object templates. We demonstrate that the correspondence information can successfully be utilised on problems that can be geometrically represented on the template shape.
Furthermore, we unify the problems of dense shape regression and articulated pose of estimation of deformable objects, by proposing the first, to the best of our knowledge, end-to-end trainable architecture that performance dense shape estimation and face landmark/body part localization. Throughout the paper, we focus on face and body shapes, where applications are abundant and benchmarks allow a fair comparison. We show that using our dense regression method out-of-the-box  outperforms a state-of-the-art semantic segmentation approach for the task of face-part segmentation, while when used as an initialisation for SDMs,  we obtain the state-of-the-art results on the challenging 300W landmark localization challenge. We demonstrate the generality of our method by performing experiments on the human body and human ear shapes. We believe that our method will find ubiquitous use, since it can be readily used for face-related tasks and can be easily integrated into many other correspondence problems.

\section*{Acknowledgement}
Alp Guler and Iasonas Kokkinos were supported by the European Horizon 2020 programme under grant agreement no 643666 (I-Support). G. Trigeorgis was supported by EPSRC DTA award at Imperial College London.  The work of S. Zafeiriou was funded by the EPSRC project EP/N007743/1 (FACER2VM).

\bibliography{egbib}

\section*{Appendix}
\appendix
\section{Network Structure for Hourglass-type DenseReg}
\begin{figure}[!h]
\centering
\includegraphics[width=\linewidth]{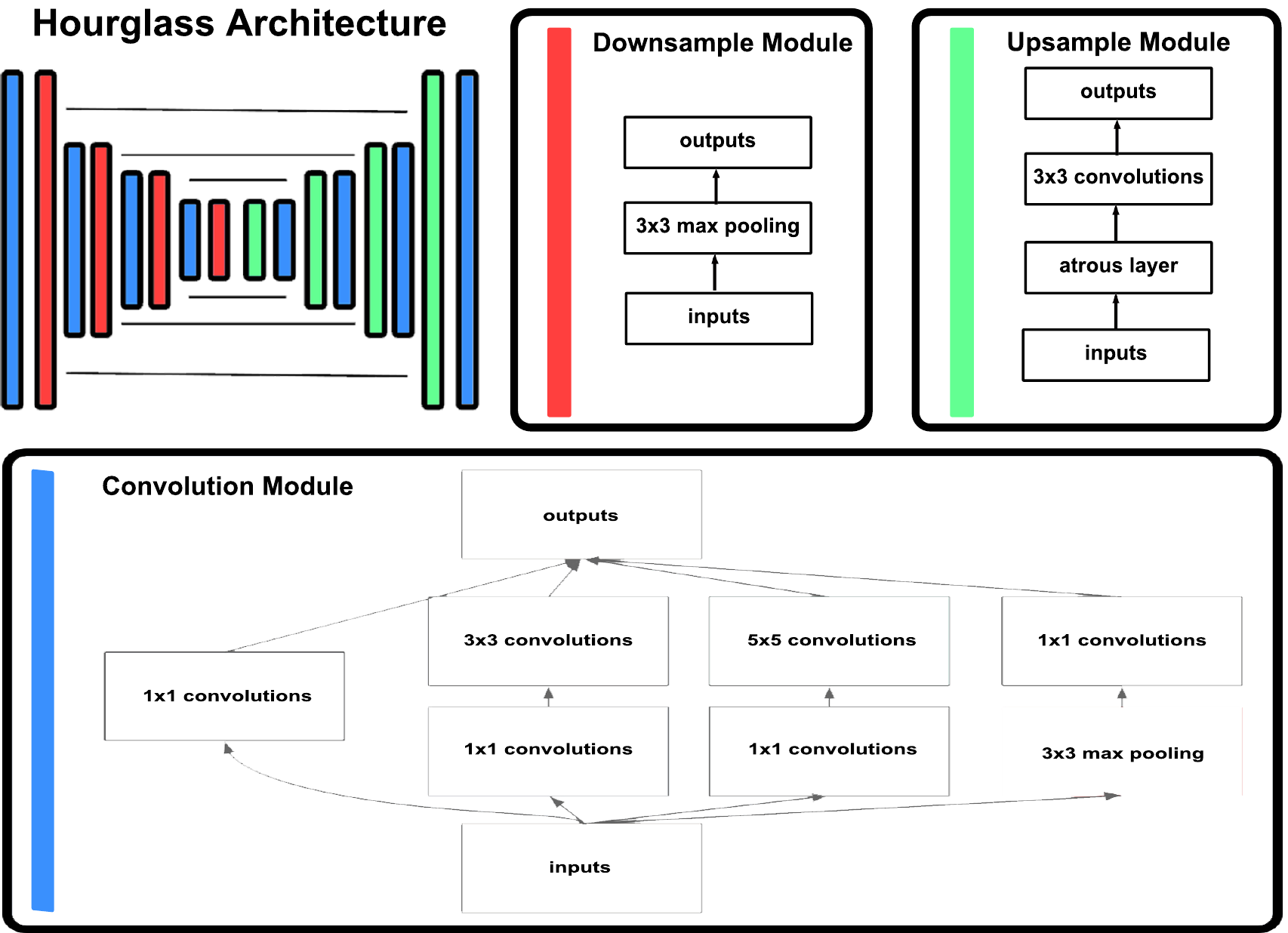}
\caption{Hourglass architecture with inception-v2 module.}
\label{fig:hourglass_arch}

\end{figure}

Figure\ref{fig:hourglass_arch} demonstrated the hourglass architecture \cite{newell2016stacked} we used with some modifications. The network is consisted by three type of modules: 1) the convolution module (blue), 2) the down sampling module (read) and 3) the up sampling module. The whole hourglass is constructed with a list convolution modules at 4 different scales with corresponding down/up sampling modules between those convolution modules. There are also bilateral connection between layers of the same scale. The composition of each type of modules are shown in the figure too. The down sampling module is just a $3\times3$ max pooling layer and the up sampling module is using a $3\times3$ atrous layer following by a $3\times3$ convolution layer. The majority of the parameters are lies in the convolution module. The original hourglass uses a chain of 3 convolution layers as its convolution module, while replacing that with the inception-v2 type module shows slight improvement on body pose estimation and obvious improvement on training speed.

\end{document}